\PassOptionsToPackage{prologue,dvipsnames,svgnames}{xcolor}
\documentclass[11pt]{article}
\usepackage{fullpage}

\usepackage{amsmath,amssymb}
\usepackage{array}
\usepackage{bm}
\makeatletter
\@ifpackageloaded{natbib}{%
  \citestyle{round}%
}{%
  \usepackage[round]{natbib}%
}
\makeatother

\usepackage{amssymb}
\usepackage{mathrsfs}
\usepackage{graphicx}
\usepackage{tabularx}
\usepackage{subcaption}
\usepackage{pdfpages}
\usepackage{afterpage}
\usepackage{caption}
\usepackage{multirow}
\usepackage{rotating}
\usepackage[flushleft]{threeparttable}
\usepackage{multirow}
\usepackage{enumitem} 
\usepackage{algorithm2e}
\usepackage{svg}
\usepackage{stackengine}
\usepackage{float}
\usepackage[most]{tcolorbox}
\RestyleAlgo{ruled}
\SetAlgorithmName{Procedure}{procedure}{List of Procedure}
\usepackage{tikz} 
\usetikzlibrary{matrix}

\usepackage{preamble/star}
\usepackage[group-separator={,},group-minimum-digits={3}]{siunitx}

\makeatletter
\@ifpackageloaded{hyperref}{%
  \hypersetup{colorlinks,linkcolor=blue,anchorcolor=blue,citecolor=blue,urlcolor=black}%
}{%
  \usepackage[colorlinks,linkcolor=blue,anchorcolor=blue,citecolor=blue,urlcolor=black]{hyperref}%
}
\makeatother
\usepackage{cleveref}

\def\T{{ \mathrm{\scriptscriptstyle T} }}
\renewcommand{\top}{\T}

\newcolumntype{P}[1]{>{\centering\arraybackslash}p{#1}}

\def\T{{ \mathrm{\scriptscriptstyle T} }}

\def\##1\#{\begin{align}#1\end{align}}
\def\$#1\${\begin{align*}#1\end{align*}}

\newcommand{\Rom}[1]{\text{\uppercase\expandafter{\romannumeral #1\relax}}}

\usepackage{color}

\definecolor{wjs}{RGB}{200,0,50}

\renewcommand{\cite}{\citep}

\AddToHook{env/tcbraster/begin}{\let\shortstack\arxivshortstack}
\makeatletter
\let\arxivendfloatbox\@endfloatbox
\renewcommand{\@endfloatbox}{%
  \arxivendfloatbox
  \ifdim\dimexpr\ht\@currbox+\dp\@currbox\relax>\textheight
    \global\setbox\@currbox\vbox{%
      \hbox to\columnwidth{\hfil
        \resizebox*{!}{0.99\textheight}{\box\@currbox}%
      \hfil}%
    }%
  \fi
}
\makeatother

\newcommand{\teaserwidth}{0.75\linewidth}
\newcommand{\sftsummarypanelwidth}{0.31\textwidth}
\newcommand{\sftsummarypanelgap}{\hspace{0.02\textwidth}}

\title{
Rethinking Synthetic Rationale Supervision for Disease Prediction
}

\author{
Buxin Su\thanks{University of Pennsylvania; Email: \texttt{subuxin@sas.upenn.edu}.} \quad
Bingxuan Li\thanks{University of Illinois Urbana-Champaign; Email: \texttt{bl61@illinois.edu}.} \quad
Cheng Qian\thanks{University of Illinois Urbana-Champaign; Email: \texttt{chengq9@illinois.edu}.} \quad
Yiwei Wang\thanks{University of California, Merced; Email: \texttt{yiweiwang2@ucmerced.edu}.} \quad
Jin Jin\thanks{University of Pennsylvania; Email: \texttt{jin.jin@pennmedicine.upenn.edu}.} \quad
Bingxin Zhao\thanks{University of Pennsylvania; Email: \texttt{bxzhao@upenn.edu}.}
}

\date{\today}

\begin{document}
\maketitle
\begin{abstract}
Supervised fine-tuning with synthetic rationales is widely assumed to improve clinical prediction by teaching models not only what to predict but why. We test this assumption on five-year Alzheimer's disease and related dementias (ADRD) prediction from longitudinal health histories. In this setting, we generate training rationales using the correct labels, then fine-tune models to output a rationale followed by a label. Across a large-scale controlled experiment of 504 configurations, we find that Rationale First SFT consistently and substantially hurts predictive performance relative to Label Only fine-tuning. The degradation persists across model families and data scales, and is not resolved by using a reasoning-oriented base model. A fixed-protocol cardiovascular disease (CVD) validation reproduces similar findings. Crucially, the failure is not explained by poor rationale quality: human expert annotation confirms that the generated rationales are medically accurate and faithfully grounded in patient-specific evidence, and few-shot experiments show that the same rationales improve performance when used as inference-time demonstrations rather than training targets.
A simple label-ordering intervention clarifies the mechanism: placing the prediction before the same rationale restores predictive performance. This suggests that conditioning predictions on generated rationales contributes to the degradation. We hope our work paves the way toward a more precise understanding of when and how rationale-based supervision helps and when it does not, guiding the responsible development of language models for high-stakes clinical prediction.
\end{abstract}

\providecommand{\teaserwidth}{\linewidth}
\begin{figure}[!t]
\centering
\includegraphics[width=\teaserwidth]{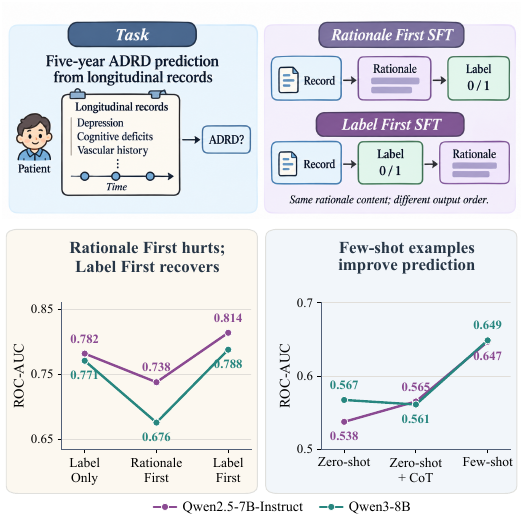}
\caption{Rationale First SFT lowers ROC-AUC relative to Label Only, while Label First recovers performance with the same rationale text. Few-shot demonstrations improve inference-time prediction in both models. The left panel shows the target-order experiment; the right averages over matched decoding settings.}
\label{fig:teaser}
\end{figure}

\section{Introduction}

\begin{figure*}[!htp]
\centering
\begin{tcbraster}[
  raster columns=3,
  raster equal height=rows,
  raster column skip=1mm
]
\begin{tcolorbox}[
  colback=blue!5,
  colframe=blue!55!black,
	  title={\shortstack[l]{Label Only}},
  fonttitle=\bfseries\scriptsize,
  boxrule=0.35pt,
  arc=1mm,
  left=1mm,
  right=1mm,
  top=1mm,
  bottom=1mm
]
{\scriptsize\raggedright
\texttt{sex}: Male\\
\texttt{medical history}:\\
F10 alcohol-related disorder: 58.4\\
G40 epilepsy: 62.3\\
F32 depressive episode: 69.3\\
BMI\_high: 70.0\\
Fluid intelligence low: 70.0\\
J06 acute upper respiratory infection: 73.6\\
I67 cerebrovascular disease: 73.1\\
E78 lipidaemias: 74.7\\[0.25em]
\texttt{ADRD}:\\
\texttt{<THINK>}\\
\texttt{</THINK>}\\
\texttt{Prediction (0 or 1): 1}
\par}
\end{tcolorbox}
\begin{tcolorbox}[
  colback=green!5,
  colframe=green!45!black,
	  title={\shortstack[l]{Free rationale (Rationale First SFT)}},
  fonttitle=\bfseries\scriptsize,
  boxrule=0.35pt,
  arc=1mm,
  left=1mm,
  right=1mm,
  top=1mm,
  bottom=1mm
]
{\scriptsize\raggedright
Same disease history.\\[0.25em]
\texttt{ADRD}:\\
\texttt{<THINK>}\\
This male has multiple mid-to-late life factors that plausibly precede
Alzheimer's or related dementia, including alcohol-related disorder,
epilepsy, low fluid intelligence, slowed matching time, cerebrovascular
disease, and later dyslipidaemia. The combination of cognitive testing
abnormalities and cerebrovascular disease in the early 70s makes an ADRD
code within the next five years plausible.\\
\texttt{</THINK>}\\
\texttt{Prediction (0 or 1): 1}
\par}
\end{tcolorbox}
\begin{tcolorbox}[
  colback=orange!7,
  colframe=orange!65!black,
	  title={\shortstack[l]{Stepwise rationale (Rationale First SFT)}},
  fonttitle=\bfseries\scriptsize,
  boxrule=0.35pt,
  arc=1mm,
  left=1mm,
  right=1mm,
  top=1mm,
  bottom=1mm
]
{\scriptsize\raggedright
Same disease history.\\[0.25em]
\texttt{ADRD}:\\
\texttt{<THINK>}\\
Step 1: Relevant ADRD-related events include I67 at 73.1, earlier G40 and
F32, and cognitive testing abnormalities at 70.0.\\
Step 2: Cerebrovascular disease in the early 70s, together with prior
cognitive deficits, makes near-term ADRD risk high.\\
Step 3: Progression to an ADRD-coded diagnosis within five years is
therefore supported.\\
\texttt{</THINK>}\\
\texttt{Prediction (0 or 1): 1}
\par}
\end{tcolorbox}
\end{tcbraster}
\caption{Example training records constructed from one participant record. The
three columns share the same disease history; only the \texttt{ADRD} response
field changes across Label Only, free-rationale, and
stepwise-rationale.}
\vspace{-0.2in}
\label{fig:data-example}
\end{figure*}

Supervised fine-tuning with synthetic rationale data has become a widely used technique for improving the reasoning capabilities of language models in medicine \citep{chen2024huatuogpt, yu2025finemedlm, kim2025small, zhuang2026towards}. The intuition is compelling: If a model learns not just what the answer is but why, it should generalize better, produce more interpretable outputs,
and be easier to audit. This intuition has motivated a growing body of work using LLM-generated rationales to improve clinical diagnosis \cite{kwon2024large}, reasoning-enhanced prediction from structured health data \cite{jiang2025reasoning,cao2026remedi}, multimodal clinical rationale generation \cite{niu-etal-2025-knowledge}, and large-scale medical reasoning datasets \cite{sun2025reasonmed,wu2025medreason}. A recurring finding is that rationale quality matters: filtering or selecting higher-quality rationales improves distillation
into smaller models \cite{song-etal-2025-rationale}, and multi-task rationale objectives can strengthen prediction alongside explanation \cite{hasan2025reason2decide} \citep[see also][]{zhuang2026chips}.

In this work, we test this intuition on five-year ADRD prediction from longitudinal health histories, a challenging task with sparse records and heterogeneous risk factors. This prediction task is clinically important and epidemiologically well motivated. Dementia is currently the seventh
leading cause of death worldwide and a major cause of disability and dependency among older adults \cite{who2025dementia}. It is a difficult
prediction target: risk can accumulate through genetic, vascular, metabolic, psychiatric, cognitive, and lifestyle pathways, rather than through one
defining precursor~\cite{reitz2023global,
rasmussen2023modifiable}. This sparsity and heterogeneity make the task a precise stress test for rationale-based SFT. Our cohort contains 42,566 participants, including 8,802 ADRD cases and 33,764 matched controls, represented with 1,167 input features. Records are sparse, with only 17.7 observed features on average. The useful signal is not a single diagnosis or a fixed checklist: one future ADRD case may have vascular and metabolic history, another may have cognitive and psychiatric signals, and another may have a different mixture of weak risk
factors. 

Across a controlled sweep of 504 configurations, models trained to output only the label (\emph{Label Only SFT}) substantially outperform models trained to output a rationale before the label (\emph{Rationale First SFT}).
Mean ROC-AUC is 0.734 for Label Only SFT, compared with 0.604 and 0.592 for the free-rationale and stepwise-rationale variants of Rationale First SFT, respectively. The gap persists across training set sizes and base models. 
The natural explanation is that the rationales are simply not good enough. We test rationale quality through two independent checks. First, few-shot experiments show that when the same style of rationale is used as a demonstration rather than a training target, it improves performance over zero-shot baselines, demonstrating that the rationales carry genuine discriminative signal. Second, human annotation confirms that the generated rationales are medically accurate and faithfully select patient-specific evidence from the record. 
We conducted further experiments and found that the order of prediction and rationale generation plays a critical role.
Switching to \emph{Label First SFT}, which trains the model to output the label before the rationale, restores predictive performance using the same free-rationale text. The problem is therefore not what rationales contain, but what happens when a model is trained to reproduce them.

We conduct further analyses to understand why examples containing rationales improve few-shot prediction, while Rationale First SFT reduces predictive performance. One possible explanation is that the two settings use rationales differently. Few-shot demonstrations provide complete examples of how to interpret patient records and make predictions, without updating model parameters. Rationale First SFT instead trains the model to reproduce explanations before predicting the label. A medically plausible explanation connects a patient’s history to their outcome by emphasizing clinically relevant risk factors. However, these factors may be common among both future cases and matched controls. Learning to reproduce such explanations therefore does not necessarily teach the model which features distinguish the two groups. This distinction is particularly relevant when prediction depends on different combinations of features across patients: an explanation can provide useful clinical guidance without fully capturing the information needed for accurate prediction.

Our additional Label First SFT study suggests that predictive performance can recover when the training sequence better matches how the rationales were generated. To generate each rationale, we provide the model with both the patient record and the correct label. The resulting explanation describes why the record is consistent with an outcome that is already known. Rationale First SFT, however, asks the model to generate this explanation before predicting the label. During training, label prediction follows an explanation written with knowledge of the correct answer; at inference, it follows the model’s own generated explanation, which may be inaccurate or misleading. Label First SFT instead trains the model to predict the label directly from the record and then generate an explanation conditioned on that label. This preserves the label information available during rationale construction and allows the prediction to be made without relying on a previously generated explanation. The recovery of predictive performance with the same rationale content is consistent with this explanation.

\section{Experimental Design}
\label{sec:experimental-design}

\subsection{Data and Task Formulation}

We study five-year Alzheimer's disease and related dementias (ADRD) prediction
from longitudinal health histories. For each participant, the input
contains prior events and risk factors available before an age-specific cutoff;
the label is whether ADRD is recorded within the next five years. Full
data-processing and matching details are in
Appendix~\ref{app:data-processing}.

\paragraph{Prediction target.}
ADRD onset is the first recorded occurrence of one of five ICD-10 code groups:
F00, F01, F03, G30, or G31. The binary label is \(Y=1\) for participants with
an ADRD onset under this definition and \(Y=0\) for controls without a recorded
ADRD onset in the processed data. The input is a history of earlier events and
risk factors aligned by age, not a static feature vector.

\paragraph{Input representation.}
Each participant is serialized as sex plus an age-indexed medical-history
dictionary, as illustrated in Figure~\ref{fig:data-example}. The final cohort
contains 42,566 participants, including 8,802 ADRD cases and 33,764 matched
controls. The structured input has 1,167 possible features: 1,102 ICD-10
first-occurrence disease features and 65 cognitive or lifestyle features.
Records are sparse, with 17.7 observed features on average and a median of 15
(interquartile range 10--23). This sparsity makes generated rationales a
demanding interface: a short explanation must select patient-specific evidence
from many weak and incomplete signals.


\subsection{Training With Labeled Targets}
\label{sec:training}
We compare SFT targets with generated rationales first (Rationale First SFT) and without rationales (Label Only SFT).
Each model is trained on the same ADRD prediction
task and the same structured health-history inputs. The only difference across the main data conditions is the response format used as the training target. We
compare Label Only, free-rationale, and stepwise-rationale targets, as shown
in Figure~\ref{fig:data-example}.
The rationales used in the free-rationale and stepwise-rationale
conditions are generated before SFT from the original training labels \citep{han2023large}. The
generator receives the structured patient record and the ground-truth
ADRD-within-five-years label, and is instructed to use only evidence present in
the record. Full generation prompts and SFT prompts are given in
Appendix~\ref{app:training}.

The controlled SFT grid crosses rationale format, training sample size,
learning rate, base model, and decoding setting
(Appendix Table~\ref{tab:controlled-experiments}). It contains 504
configurations: three target formats, three sample sizes, four learning rates,
two base models, and seven decoding settings. The base models are Qwen3-8B and
Qwen2.5-7B-Instruct. The decoding settings include greedy decoding and top-\(k\)
or top-\(p\) sampling at temperatures 0.1, 0.5, and 1.0. Every configuration is
evaluated on the same held-out test set of 853 individuals.
For matched comparisons, we vary one factor at a time and hold the remaining
factors fixed. We use paired \(t\)-tests for grid-average comparisons and a
paired DeLong test for selected ROC-AUC comparisons between fixed
configurations, following Appendix~\ref{app:metric}.






\section{Experimental Results}

This section evaluates whether the SFT target should include a generated
rationale before the final ADRD risk probability. Across the controlled sweep,
models trained to output only the final label or probability achieve higher
ROC-AUC than models trained to output either free-form or stepwise rationales
before the prediction. This pattern remains when we increase the training set size and
when we use Qwen3-8B, a reasoning-oriented base model. We then report three
additional checks from the SFT sweep: how performance changes with training-set
size, how the two base models behave under each target format, and how decoding
choices affect the trained models. Finally, two fixed experiments assess
cross-disease transfer and target ordering.

\subsection{Label Only Outperforms Rationale First SFT}

\providecommand{\sftsummarypanelwidth}{0.44\textwidth}
\providecommand{\sftsummarypanelgap}{\hfill}
\begin{figure}[!t]
\centering
\begin{subfigure}[t]{\sftsummarypanelwidth}
\centering
\includegraphics[height=0.22\textheight,keepaspectratio]{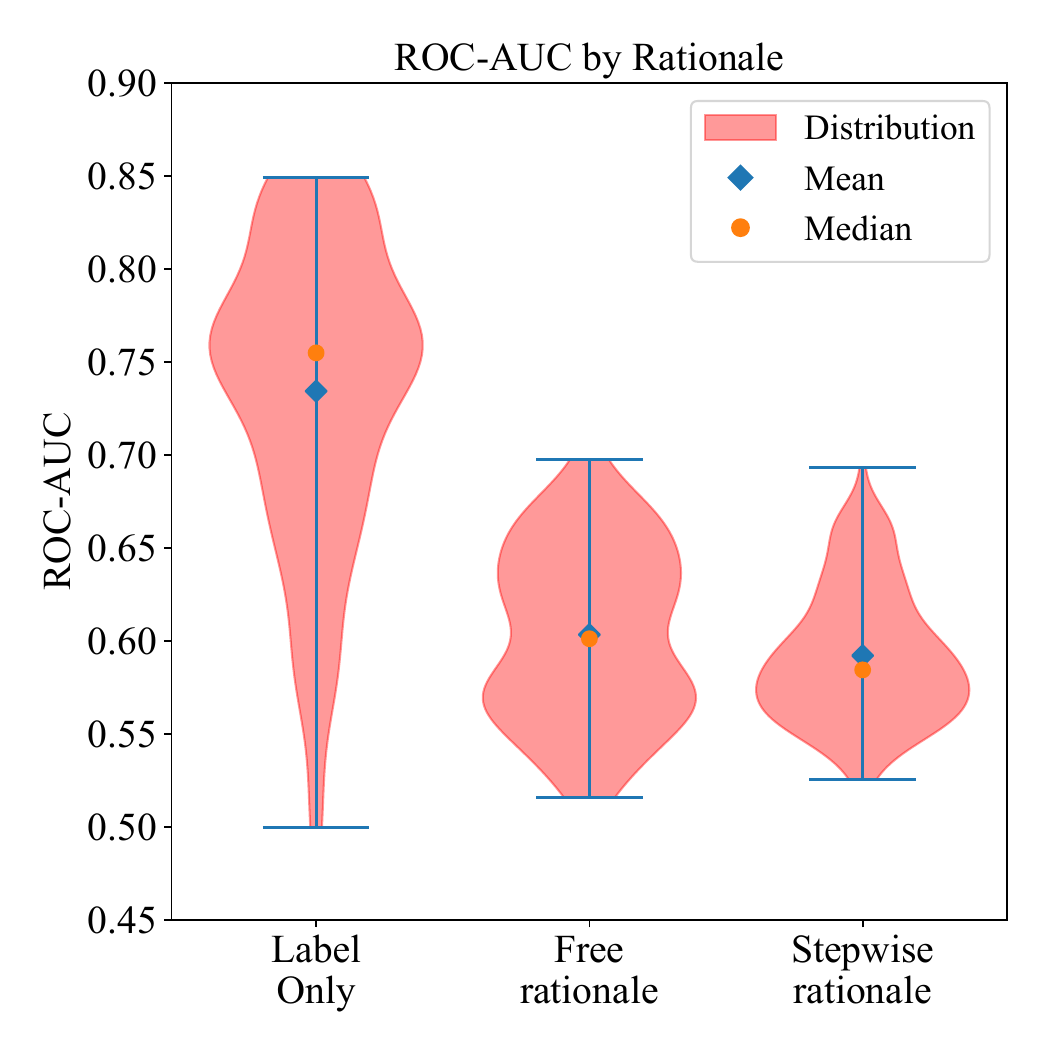}
\caption{Rationale, ROC-AUC}
\end{subfigure}
\sftsummarypanelgap
\begin{subfigure}[t]{\sftsummarypanelwidth}
\centering
\includegraphics[height=0.22\textheight,keepaspectratio]{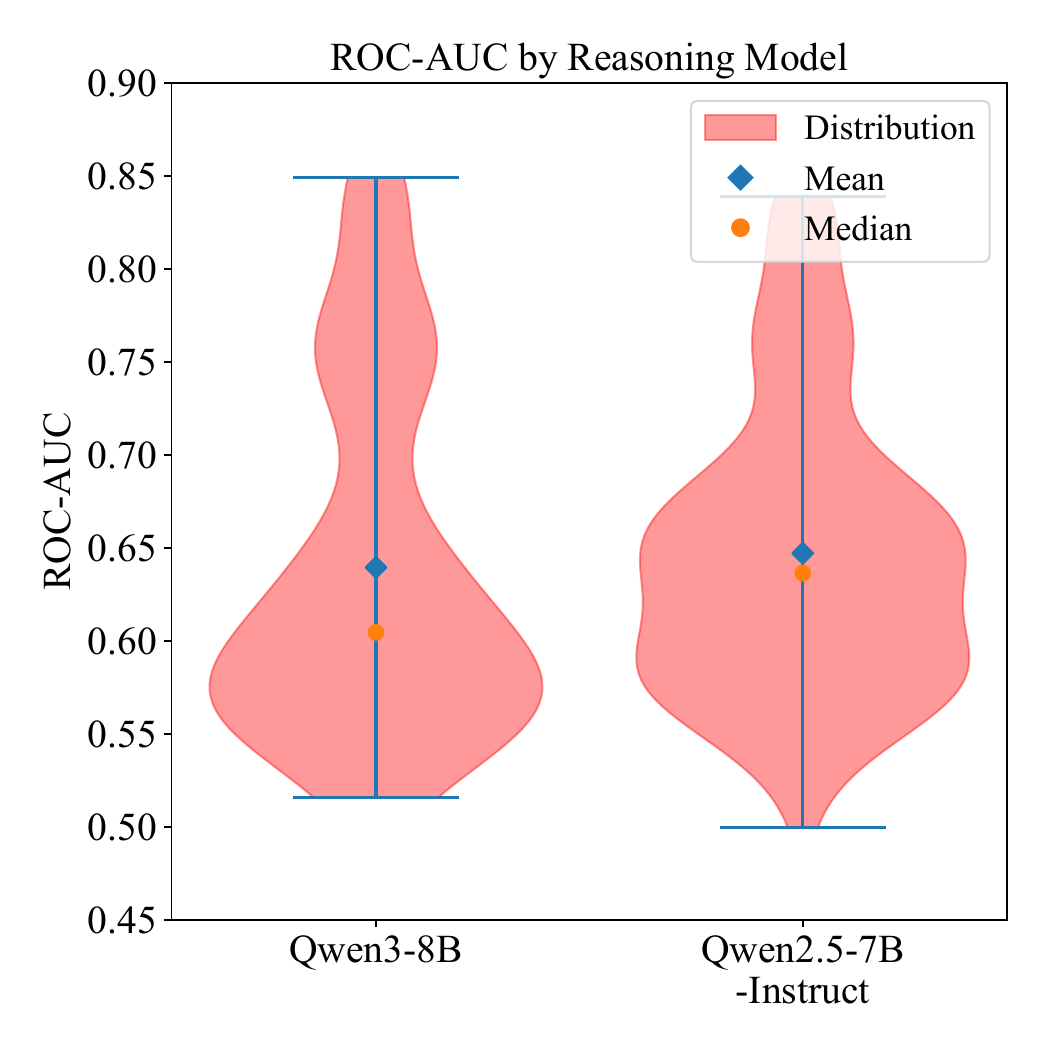}
\caption{Base model, ROC-AUC}
\end{subfigure}
\caption{SFT ROC-AUC performance by rationale format and base model.}
\vspace{-0.3in}
\label{fig:sft-rationale-ranking}
\end{figure}

\begin{figure*}[!tp]
\centering
\includegraphics[width=\textwidth]{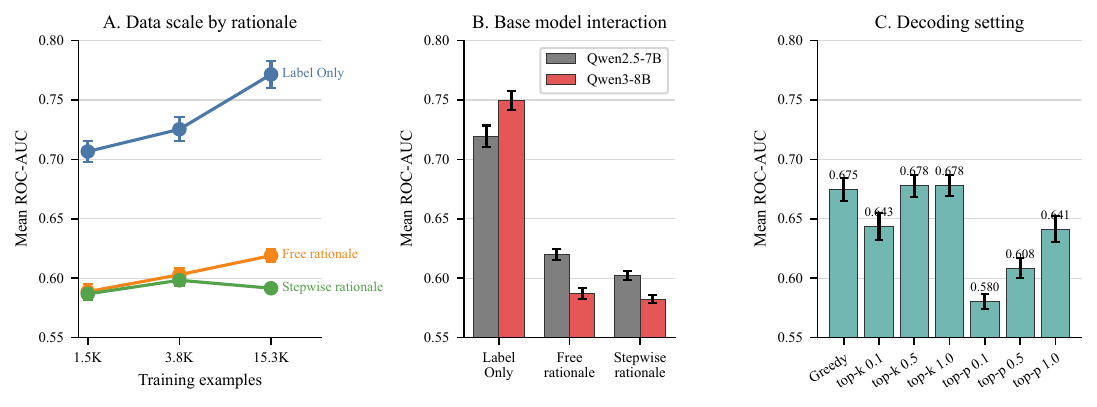}
\vspace{-0.2in}
\caption{Parameter-level SFT diagnostics for the additional summary insights. All
panels use ROC-AUC as the main metric and summarize the full SFT sweep. Bars or
points show mean ROC-AUC, and error bars show standard errors across configurations. Panel
A shows data scaling by rationale format, Panel B shows the base-model
interaction with target format, and Panel C compares decoding settings.}
\vspace{-0.2in}
\label{fig:sft-parameter-insights}
\end{figure*}

\paragraph{Direct label targets without rationales are strongest.}
Across matched SFT configurations, Label Only is clearly strongest
(Figure~\ref{fig:sft-rationale-ranking}A). Mean ROC-AUC is 0.734 for
Label Only, compared with 0.604 for free-rationale and
0.592 for stepwise-rationale. Both rationale conditions
are substantially worse than Label Only (paired \(t\)-test,
\(P=7.26\times10^{-52}\) for free-rationale and
\(P=6.51\times10^{-57}\) for stepwise-rationale).

The same pattern appears beyond ROC-AUC (Appendix
Figure~\ref{fig:app-sft-rationale-pr}C). Label Only has
higher mean PR-AUC than free-rationale (0.504 versus 0.313;
\(P=1.79\times10^{-49}\)) and stepwise-rationale (0.504 versus
0.306; \(P=1.92\times10^{-52}\)).

\paragraph{The gap remains after selecting the best configuration.}
The best individual SFT configurations lead to the same conclusion. The best
Label Only configuration reaches ROC-AUC 0.849, while the best
free-rationale and stepwise-rationale configurations reach ROC-AUC 0.698 and
0.693. Because the Label Only median is 0.755, even the best observed
rationale configurations fall below a typical Label Only configuration.



\paragraph{A reasoning-oriented base model does not eliminate the degradation.}

Using a reasoning-oriented base model does not remove the rationale degradation.
Across 252 matched SFT pairs, Qwen3-8B has slightly lower mean ROC-AUC than
Qwen2.5-7B-Instruct (0.640 versus 0.647; mean difference \(-0.0077\); paired
\(t\)-test, \(P=0.0348\); Figure~\ref{fig:sft-rationale-ranking}B). The absolute
difference is small, but it goes in the opposite direction from the idea that a
reasoning-oriented base model should handle rationales better.

The best-configuration comparison is also not significant. The best Qwen3-8B
SFT configuration reaches ROC-AUC 0.849, compared with 0.839 for the best
Qwen2.5-7B-Instruct configuration
(paired DeLong test, \(P=0.235\)). Appendix
Figure~\ref{fig:app-sft-model-pr} shows the same lack of a practical advantage
on PR-AUC, F1, and recall. Thus, the SFT degradation is not resolved by switching from
a standard instruction model to a model with a stronger reasoning emphasis.

\subsection{Analysis and Insights}

\begin{figure*}[!t]
\centering
\includegraphics[width=0.94\textwidth]{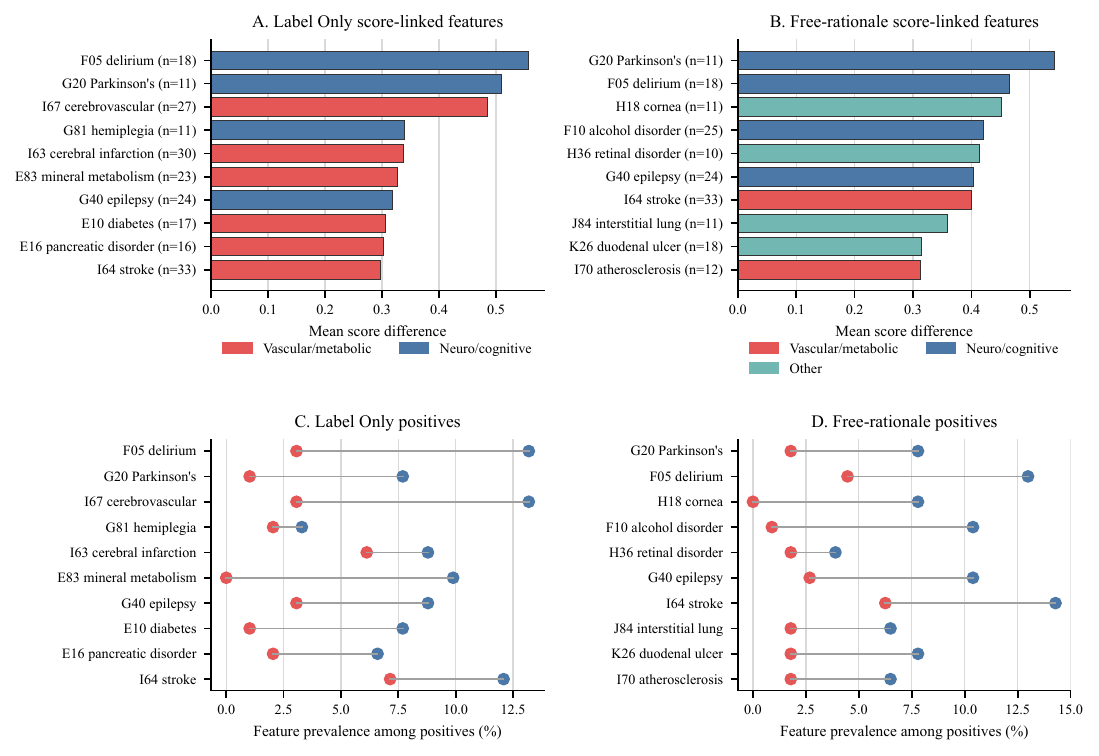}
\vspace{-0.1in}
\caption{Feature-level error analysis for the best Label Only and
free-rationale SFT configurations, using the same test examples and the
same mapped feature representation. Panels A and B show post-hoc feature-score associations
for the top mapped features in the Label Only and free-rationale
configurations, respectively. Panels C and D compare the prevalence of the same features among
detected positives (blue) and missed positives (red). In Panels A--B, $n$ counts
participants with each feature in the full test set; score differences use
available scores. 
}
\vspace{-0.2in}
\label{fig:free-rationale-error-feature}
\end{figure*}

\paragraph{Data scaling benefits Label Only targets most.}
Increasing the sample size improves Label Only SFT monotonically. Mean ROC-AUC rises from
0.707 at 1.5K examples to 0.725 at 3.8K and 0.771 at 15.3K; in matched
comparisons, 15.3K improves over 1.5K by \(+0.065\) ROC-AUC on average
(\(P=7.19\times10^{-10}\)). PR-AUC also rises from 0.451 to 0.568.
Scaling is weaker for rationale targets: free-rationale
improves by only \(+0.030\) ROC-AUC from 1.5K to 15.3K, and stepwise-rationale
has no significant ROC-AUC gain (\(+0.0048\), \(P=0.337\);
Figure~\ref{fig:sft-parameter-insights}A).

\paragraph{Model choice interacts with rationale format.}
Qwen3-8B improves performance when the target contains only the label, but
degrades performance when the target contains generated rationales. Under
Label Only SFT, Qwen3-8B has
mean ROC-AUC 0.750 versus 0.719 for Qwen2.5-7B-Instruct
(matched \(+0.030\), \(P=3.69\times10^{-4}\)). Under free-rationale SFT,
Qwen3-8B drops to 0.587 versus 0.620 for Qwen2.5-7B-Instruct
(\(-0.033\), \(P=9.91\times10^{-14}\)); under stepwise-rationale SFT, it is
0.582 versus 0.602 (\(-0.020\), \(P=5.54\times10^{-8}\)). PR-AUC validates
the same interaction: Qwen3-8B is higher for Label Only (0.529 versus
0.479), but lower for free-rationale (0.297 versus 0.328) and
stepwise-rationale (0.296 versus 0.316; Figure~\ref{fig:sft-parameter-insights}B).

\paragraph{Decoding choice is a strong inference-time parameter.}
The inference setting changes ROC-AUC substantially even after SFT. Averaged
over all training settings, top-\(p\) sampling has lower mean ROC-AUC than
top-\(k\) sampling or greedy decoding (0.610 versus 0.666 and 0.675). The
effect is clearest for Label Only SFT: greedy and all top-\(k\) temperatures
cluster near ROC-AUC 0.768, while top-\(p\) at temperature 0.1 falls to 0.615
and top-\(p\) at temperature 0.5 reaches only 0.691. PR-AUC supports the same
interpretation, with Label Only top-\(p\) at temperature 0.1 reaching 0.394
compared with roughly 0.531 for greedy or top-\(k\) decoding. Thus, stable
decoding is especially important for preserving the gains from direct
Label Only fine-tuning (Figure~\ref{fig:sft-parameter-insights}C).

\paragraph{Label First target design.}
We further test whether the ADRD degradation is caused by rationale supervision
itself or by target order. We compare Label Only SFT, the original rationale
$\rightarrow$ prediction target, and a prediction $\rightarrow$ rationale target containing the same free-rationale text. The respective ROC-AUC/PR-AUC values are 0.782/0.473, 0.738/0.390, and 0.814/0.567 for Qwen2.5-7B-Instruct, and 0.771/0.516,
0.676/0.404, and 0.788/0.505 for Qwen3-8B. Label First ordering clearly
outperforms Rationale First SFT on both models and matches Label Only
SFT. These results suggest that synthetic rationale supervision is not
inherently harmful in this setting. Because the rationales were generated with
access to the correct label, placing the label first better matches their
construction and allows prediction directly from the record, without relying
on a previously generated explanation. The recovery of predictive performance
with the same rationale content is consistent with this explanation.

\paragraph{Cross-disease validation.}
To assess whether the primary ADRD result extends beyond its original endpoint,
we conduct a fixed-protocol comparison on five-year cardiovascular disease
(CVD) prediction, defined by the first recorded occurrence of ICD-10 codes I20--I25 or I60--I64 with a
one-year prediction offset. Label Only and Rationale First free-rationale
Qwen3-8B SFT use the same CVD records (1,533 training; 853 held out), seed 42,
three epochs, a learning rate of $1.5\times10^{-4}$, and greedy decoding.
Label Only/free-rationale SFT yields ROC-AUC 0.669/0.632 and PR-AUC
0.344/0.302, giving the same direction as ADRD for both ranking metrics without
retuning.

\section{Discussion}

We examine why Rationale First SFT degrades prediction through expert review,
few-shot demonstrations, error analysis, and target-order experiments.

\subsection{Quality Analysis of Generated Rationales}

\subsubsection{Human Expert Study}
\label{sec:human_eval}

\begin{table}[!htp]
\centering
\label{tab:adrd_rationale_quality}
\footnotesize
\setlength{\tabcolsep}{6pt}
\renewcommand{\arraystretch}{1.12}
\begin{tabular}{@{}lcccc@{}}
\toprule
\textbf{Sex} &
\textbf{Logic} &
\textbf{Bio.} &
\textbf{ADRD} &
\textbf{Fidelity} \\
\midrule
Male   & 8.67 & 8.33 & 8.67 & 7.83 \\
Female & 8.00 & 7.00 & 8.00 & 7.00 \\
\bottomrule
\end{tabular}
\caption{Logic = logical coherence; Bio.\ = biomedical correctness; ADRD = ADRD relatedness; Fidelity = evidence fidelity and temporal grounding.}
\vspace{-0.1in}
\label{tab:human}
\end{table}

To test whether low-quality generation causes the degradation, a clinical
expert evaluated a stratified sample of training rationales for \emph{medical
accuracy}---factual correctness and consistency with established ADRD risk
factors---and \emph{record fidelity}---use of only evidence in the structured
record, without invented symptoms, medications, or diagnoses. Ratings were high
on both dimensions: the rationales correctly interpreted recorded ICD-10 codes,
invoked established ADRD pathways, and avoided fabricated patient evidence.
Table~\ref{tab:human} shows broadly comparable averages across sexes; male
samples scored slightly higher on biomedical correctness (8.33 vs.\ 7.00) and
fidelity (7.83 vs.\ 7.00). Thus, by clinical
standards, the rationales are high quality rather than too noisy or inaccurate
for training.

\subsubsection{Rationales as Few-Shot Demonstrations}
\label{sec:fewshot-ablation}

\begin{figure*}[!t]
\centering
\includegraphics[width=\textwidth]{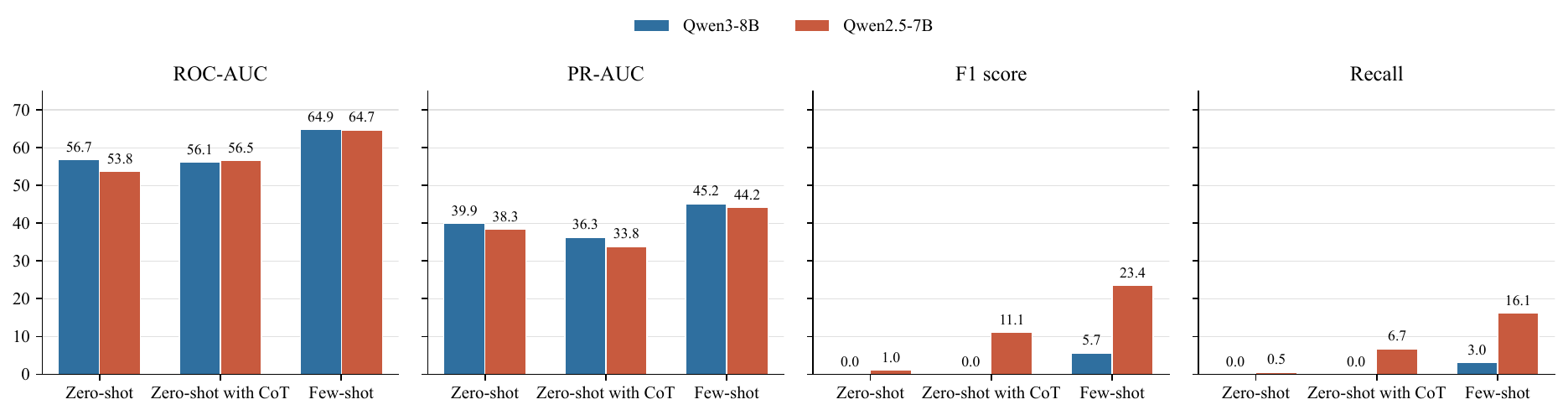}
\caption{Few-shot ablation in the training-free setting. Bars show metric means
over matched decoding settings within each model, displayed as percentages.
The three conditions are Zero-shot, Zero-shot with CoT, and Few-shot.}
\vspace{-0.2in}
\label{tab:fewshot-ablation}
\end{figure*}

A second check asks whether rationales that fail as training targets still help
as demonstrations. We prepend five fixed GPT-5.4 disease-prediction examples,
each pairing a structured history with a \texttt{<THINK>} block and final
probability, to the query prompt. This adds worked examples to the Zero-shot
with CoT baseline without updating model parameters.

Few-shot demonstrations substantially improve both zero-shot baselines. Across
models and matched decoding settings, mean ROC-AUC rises from 0.552 (Zero-shot)
and 0.563 (Zero-shot with CoT) to 0.648 (Few-shot; paired $t$-tests,
$P=6.89\times10^{-9}$ and $P=7.57\times10^{-7}$, respectively;
Figure~\ref{tab:fewshot-ablation}). Deterministic Qwen3-8B likewise rises from
0.510 to 0.654. Few-shot scores reach 0.85 and yield three true positives at
the 0.5 threshold; Zero-shot with CoT never exceeds 0.35 and predicts no
positives.

\begin{figure*}[!t]
\centering
\includegraphics[width=\textwidth]{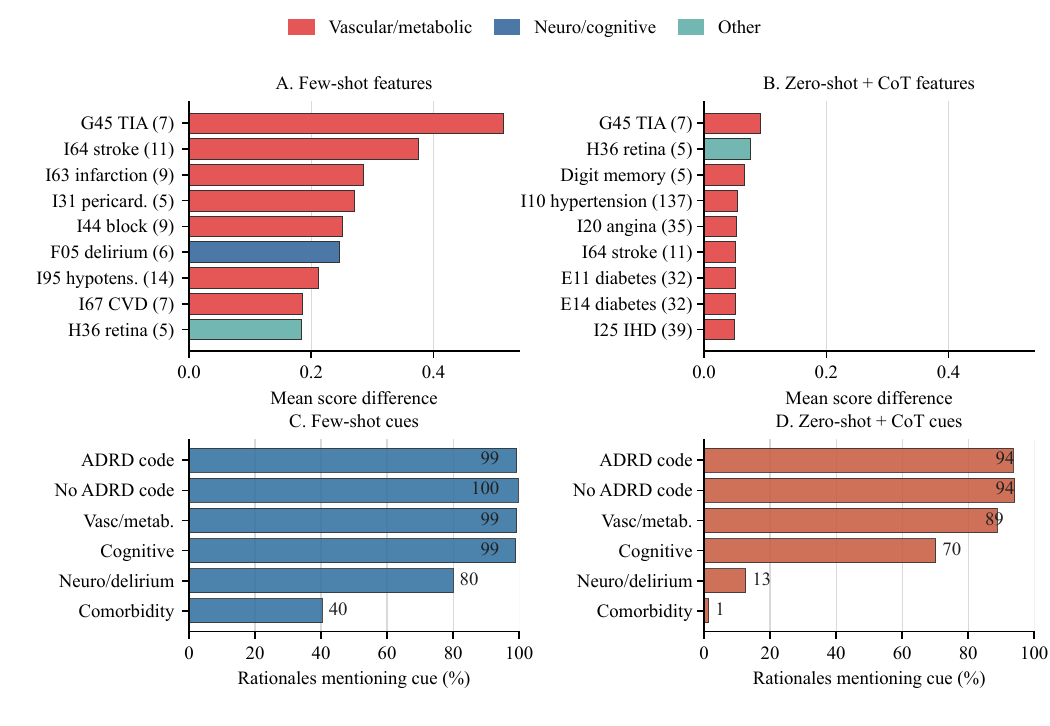}
\vspace{-12mm}
\caption{Deterministic Qwen3-8B few-shot and Zero-shot with CoT analysis. Panels
A--B show mapped-feature score associations (mean score with versus without a
feature; minimum five examples), as in
Figure~\ref{fig:free-rationale-error-feature}. Panels C--D show keyword-matched
rationale-cue frequencies in generated \texttt{<THINK>} text.}
\vspace{-0.2in}
\label{fig:fewshot-feature-reasoning}
\end{figure*}

Feature analysis clarifies the gain (Figure~\ref{fig:fewshot-feature-reasoning}).
Few-shot scores rise most for specific ADRD-related histories---transient
cerebral ischaemia, stroke, cerebral infarction, delirium, hypotension, and
cerebrovascular disease---whereas Zero-shot with CoT emphasizes broad
cardiovascular and metabolic markers such as hypertension, angina, diabetes,
and chronic ischaemic heart disease, with a maximum feature-score increase of
only 0.092 (Panels A--B). Both formats check for absent ADRD codes at similar
rates; reasoning specificity changes instead. Few-shot mentions cognitive-test
evidence in 99\% versus 70\% of cases, neurologic or delirium signals in 80\%
versus 13\%, and comorbidity caveats in 40\% versus 1\% (Panels C--D). The
demonstrations therefore connect broad risks to specific signals and separate
background comorbidity from ADRD evidence. We interpret this result as
supporting evidence that worked examples containing generated rationales can
provide useful inference-time guidance. The target-order intervention, which
directly manipulates the prediction--rationale interface, provides the more
specific test of the proposed mechanism.

\subsection{Error Analysis Across Rationale Formats}
\label{sec:error-analysis}

We next compare the best Label Only and free-rationale configurations on the
same 853 test examples. For features appearing at least ten times, a
conservative post-hoc importance proxy is the mean score when a feature is
present minus the mean when it is absent.

The Label Only model learns a tight clinical signal: specific neurologic,
vascular, and metabolic histories separate future cases from matched controls.
F05 delirium raises the mean score by 0.556 (83.3\% observed case rate), G20
Parkinson's disease by 0.510 (72.7\%), and I67 cerebrovascular disease by 0.485
(58.3\%). Other top features---hemiplegia, cerebral infarction, mineral
metabolism disorders, epilepsy, diabetes, and stroke---are likewise
disproportionately represented among future cases
(Figure~\ref{fig:free-rationale-error-feature}A,C).

The free-rationale model retains some ADRD-specific markers---Parkinson's
disease, delirium, epilepsy, and stroke---but broadens its top features to H18
corneal disorders, H36 retinal disorders, F10 alcohol-related disorders, J84
interstitial lung disease, and K26 duodenal ulcer
(Figure~\ref{fig:free-rationale-error-feature}B). These reflect general health
burden more than near-term ADRD risk. They still carry signal and occur more
often among detected than missed positives (Panel D), but overweighting general
morbidity inflates false positives among controls.

The errors make this shift concrete. Compared with Label Only SFT, free-rationale SFT has
more false positives (86 vs.\ 26) and false negatives (112 vs.\ 98), with lower
ROC-AUC (0.698 vs.\ 0.849), PR-AUC (0.384 vs.\ 0.702), precision (0.472 vs.\
0.778), and specificity (0.870 vs.\ 0.961). Of the 119 examples predicted correctly only under
Label Only, 77 are controls that free-rationale marks positive by treating
weak cognitive bins, high BMI, smoking, and dyslipidaemia as sufficient
evidence. The other 42 are cases marked negative despite hypertension,
diabetes, depression, B12 anaemia, and cognitive-timing signals because no
explicit dementia code appears. Both patterns indicate imitation of rationale
surface logic rather than the case--control boundary.

\subsection{Why Does Rationale First SFT Degrade Model Performance?}
\label{sec:why-rationales-fail}

\paragraph{Demonstrations and SFT use rationales differently.}
Few-shot demonstrations supply complete examples of how to interpret a patient
record and make a prediction, without updating model parameters. Rationale
First SFT instead trains the model to reproduce an explanation before
predicting the label. A medically plausible explanation connects a patient's
history to a known outcome, but the risk factors it emphasizes may occur in
both future cases and matched controls. Learning to reproduce the explanation
therefore need not teach the model which features distinguish these groups~\citep{ross-etal-2022-self},
particularly when prediction depends on different combinations of features
across patients. The error analysis is consistent with this distinction: the
free-rationale variant of Rationale First SFT gives greater weight to broad
morbidity markers, including retinal disorders, interstitial lung disease, and
duodenal ulcer (Section~\ref{sec:error-analysis}).

\paragraph{Rationale construction and target order.}
Let $X$ denote the patient record, $y$ the correct label, and $R(X,y)$ the
training rationale generated using both. This rationale explains an outcome
already known to the generator. For input $X$, Rationale First SFT uses the
training target $X \rightarrow R(X,y) \rightarrow y$.
During training, label prediction therefore follows a rationale
written with knowledge of the correct label $y$. At inference, however, $y$ is
unknown: the model first generates its own rationale from the
record and then predicts the label. The prediction thus depends on a generated
explanation that may be inaccurate or misleading.

Label First SFT instead uses the training target $X \rightarrow y \rightarrow R(X,y)$~\citep{chen2025distilling}. At inference, it first predicts the label directly from $X$ and then generates an explanation conditioned on $X$ and the predicted label. This better matches rationale construction, in which the label is available before the explanation is generated, and avoids making the prediction depend on a preceding explanation.
Predictive performance recovers with the same rationale text, consistent with this explanation of the role of target order.



\section{Conclusion}
We studied whether supervised fine-tuning on synthetic rationales improves
five-year ADRD prediction. Across 504 controlled configurations,
Rationale First SFT consistently underperforms Label Only SFT. Expert review
confirms that the rationales are medically accurate and grounded, while
few-shot experiments show that examples containing rationales improve prediction without updating model parameters. Learning to reproduce these explanations, however, does not necessarily teach the model which features distinguish future cases from matched controls. 
Furthermore, the training rationales were generated with the correct label, whereas Rationale First SFT requires an explanation before label prediction. Label First SFT better matches rationale construction
and allows prediction directly from the record, without a preceding
explanation. It restores predictive performance with the same rationale text. 

Although the fixed CVD comparison extends the ranking-metric direction beyond
ADRD, the comprehensive sweep and target-order intervention remain
ADRD-specific. The CVD check uses one Qwen3-8B configuration and one training
seed, and both endpoints come from the same source population; broader
cross-disease, cross-cohort, and model-family generality remains unresolved.
The study evaluates 7--8B open models and LLM-generated rather than
clinician-written rationales. Because a five-year outcome label does not define
a unique patient-level explanation, the expert audit evaluates clinical
coherence and record grounding rather than equivalence to a human-written gold
rationale. Future work should test larger models, clinician-curated evidence,
label-weighted or hybrid objectives, and diseases with stronger precursors or
denser records \citep{li2026calibrate}. The task is for research evaluation only and is not intended
for clinical deployment. These results identify a Rationale First failure mode
in this setting.

{\small
\bibliographystyle{abbrvnat}
\bibliography{ref}
}

\clearpage
\appendix

\section{Additional Details on Data Preliminaries}
\label{app:data-preliminaries}

\subsection{Data Processing}
\label{app:data-processing}

This subsection records the data-processing choices used to construct the
longitudinal ADRD prediction cohort. The pipeline starts from participant-level
first-occurrence disease tables and structured UK Biobank-derived sources \citep{bycroft2018uk}:
ADRD case and control tables, cancer first-occurrence records, cognitive test
factors, lifestyle and environmental factors, APOE genotypes, and ADRD
polygenic risk scores. Participant identifiers are cleaned before merging. For
genotype and polygenic risk score sources, identifiers are mapped to the common
participant index used across the linked UK Biobank release.

\paragraph{Feature harmonization.}
The lifestyle and environment table is converted into age-stamped categorical
features. For each group of assessment columns, entries are coerced to numeric
values; negative values and values in \([0.1,0.9]\) are set to missing. The
remaining values are pooled across assessment instances and discretized at the
empirical 10th and 90th percentiles. The middle category is removed for these
features, leaving lower-tail and upper-tail indicators. One-hot indicators are
then assigned the participant's age at the corresponding UK Biobank assessment
instance when an instance marker is present. Field names are mapped to
human-readable titles using the accompanying UK Biobank metadata.

Genetic features are processed separately before being merged into the
lifestyle table. APOE columns are collapsed into carrier indicators
(\texttt{e2\_carrier}, \texttt{e3\_carrier}, and \texttt{e4\_carrier}). The
ADRD polygenic risk score is binned by
quantiles with cut points \((0,0.10,0.30,0.60,0.90,1.00)\), producing five
ordered risk categories. APOE and polygenic risk score categories are one-hot
encoded, with active indicators assigned age 0 and inactive indicators left
missing. The lifestyle table is the anchor table, and APOE and polygenic risk
score features are merged by cleaned participant ID.

\paragraph{Case construction.}
The ADRD case table is merged with cancer first-occurrence features, cognitive
test features, and the merged lifestyle/genetic feature table on the
intersection of available participant IDs. The target column is normalized to
\texttt{adrd} and treated as the age at first ADRD onset. Feature columns are
converted to numeric values. Any feature timestamp later than
\(\texttt{adrd}-1\) is removed, giving a one-year offset between the last
eligible record and diagnosis. Cases are kept only if they have at least one
retained feature in \((\texttt{adrd}-5,\texttt{adrd}-1]\) and if ADRD onset is
at age 50 or older. The resulting case table is carried forward to matching and
serialization.

\paragraph{Matched control construction.}
Controls are drawn from participants without an ADRD onset after merging the
same feature sources on the intersection of available participant IDs. The
control table is aligned to the case schema and assigned a missing
\texttt{adrd} value. We match controls to cases on sex and observation-age
distributions, following the matched disease-prediction design of
\citet{shmatko2025learning}. The purpose of the matching is to make the label
depend on pre-diagnostic health history rather than on follow-up length. Without
this step, a model could partially separate cases from controls by the endpoint
of the observed record.

For cases, the observation cutoff is tied to the onset age. All features after
one year before ADRD onset are removed, preventing the model from using
diagnostic or near-diagnostic information. We keep a case only if at least one
retained feature lies in the interval from five years to one year before onset,
and we include only cases with onset at age 50 or older. Thus, a positive example asks the
model to predict ADRD within five years from the observed history while leaving
at least a one-year gap before the recorded onset.

For controls, we compute each case's latest retained feature age and form a
histogram using five-year age bins. For each bin, we select up to four controls
per case among controls with the same sex and at least one feature timestamp in
that bin. A cutoff age is drawn, with random seed 42, from observed control
feature ages in the bin; features after that cutoff are removed. Selected
controls are removed from the pool before later bins are processed. This
procedure yields the 1:4 case-control structure used in the experiments while
keeping the observed age distribution comparable between cases and controls.

\paragraph{Train/test split and serialization.}
Cases and controls are shuffled and split separately into 90\% training and 10\%
test partitions, using seed 42 for cases and seed 123 for controls.
Column names are mapped to full labels when a matching label is available, and
duplicate names introduced by this mapping are disambiguated. Each row is
serialized with three fields: \texttt{sex}, an ordered
\texttt{medical history} dictionary of non-missing event-age pairs sorted by
age, and the binary \texttt{ADRD} label. Rows with finite \texttt{adrd} values
receive label 1; rows with missing \texttt{adrd} values receive label 0. As a
final quality filter, examples with fewer than three medical-history entries
are removed from the training and test partitions. This rule is applied
consistently across the serialized examples used in the fine-tuning experiments.

\section{Additional Details on Metrics}
\label{app:metric}

This section formalizes the evaluation metrics and hypothesis tests used in
the controlled experiments. We consider a binary
prediction problem with test examples $\{(Y_i,S_i)\}_{i=1}^n$, where
$Y_i\in\{0,1\}$ is the observed label and $S_i\in\mathbb{R}$ is a scalar model
score, with larger scores indicating higher predicted risk. Let
$\mathcal{P}=\{i:Y_i=1\}$ and $\mathcal{N}=\{i:Y_i=0\}$ denote the empirical
positive and negative sets, with $m=|\mathcal{P}|$ and
$n_0=|\mathcal{N}|$. For a threshold $\tau$, the induced classifier is
$\widehat{Y}_i(\tau)=\mathbb{I}\{S_i\ge \tau\}$. Zero-shot and few-shot
evaluations use a fixed probability threshold of $\tau=0.5$ for F1 score and recall.

\subsection{Evaluation Metrics}

We report ROC-AUC, F1 score, and recall using their standard definitions.
ROC-AUC is the threshold-free ranking metric
$\Pr(S^+>S^-)+\frac{1}{2}\Pr(S^+=S^-)$, estimated by the usual empirical
Mann--Whitney statistic over positive--negative score pairs
\citep{hanley1982meaning,bamber1975area}. F1 is the harmonic mean of precision and
recall, following the standard F-measure formulation
\citep{vanrijsbergen1979information}, equivalently
$2\mathrm{TP}/(2\mathrm{TP}+\mathrm{FP}+\mathrm{FN})$,
and recall is
$\mathrm{TP}/(\mathrm{TP}+\mathrm{FN})$. These metrics are
included to summarize global discrimination, the precision--recall
trade-off, and sensitivity to positive cases, respectively
\citep{fawcett2006introduction}.

\paragraph{PR-AUC.}
For threshold $\tau$, define
\[
\begin{aligned}
\widehat{P}(\tau)
&=
\frac{\mathrm{TP}(\tau)}
{\mathrm{TP}(\tau)+\mathrm{FP}(\tau)},\\
\widehat{R}(\tau)
&=
\frac{\mathrm{TP}(\tau)}
{\mathrm{TP}(\tau)+\mathrm{FN}(\tau)}.
\end{aligned}
\]
The precision--recall area under the curve (PR-AUC) is the area under
precision as a function of recall as the threshold is swept over score values.
Empirically, after sorting examples by decreasing score and evaluating the
right-continuous step curve at the distinct recall values
$0=r_0<r_1<\cdots<r_K$, we use
\[
\widehat{\mathrm{PR\text{-}AUC}}
=\sum_{k=1}^{K}(r_k-r_{k-1})\,p_k,
\]
where $p_k$ is the precision after including all predictions up to the
$k$th recall level \citep{davis2006relationship}. PR-AUC is
prevalence-sensitive and is particularly informative when the positive class
is rare, since it directly penalizes false positives among high-scoring
predictions \citep{saito2015precision}.

\subsection{Hypothesis Testing}

For each controlled comparison, the null hypothesis is equality of the
population-level metric between two experimental conditions:
\[
H_0:\Delta=\theta_a-\theta_b=0,
\qquad
H_1:\Delta\ne 0.
\]
All tests are paired because every condition in the grid is evaluated on the
same held-out examples, with all non-target experimental factors controlled.

\paragraph{Paired DeLong test for ROC-AUC.}
For ROC-AUC, let
$\widehat{\Delta}=\widehat{\mathrm{AUC}}_a-\widehat{\mathrm{AUC}}_b$.
Because ROC-AUC is a two-sample U-statistic, the paired DeLong test estimates
the covariance of the two AUC estimates from their shared positive and negative
examples \citep{delong1988comparing}. Define
\[
\phi(s_i,s_j)
=\mathbb{I}(s_i>s_j)+\frac{1}{2}\mathbb{I}(s_i=s_j).
\]
For condition $r\in\{a,b\}$, the positive- and negative-class placement values
are
\[
\begin{aligned}
V_{r,i}^{(1)}
&=
\frac{1}{n_0}\sum_{j\in\mathcal{N}}\phi(S_{r,i},S_{r,j}),
&& i\in\mathcal{P},\\
V_{r,j}^{(0)}
&=
\frac{1}{m}\sum_{i\in\mathcal{P}}\phi(S_{r,i},S_{r,j}),
&& j\in\mathcal{N}.
\end{aligned}
\]
Let $\widehat{\theta}=(\widehat{\mathrm{AUC}}_a,
\widehat{\mathrm{AUC}}_b)^\top$ and estimate the paired DeLong covariance
matrix by
\[
\begin{aligned}
\widehat{\Sigma}_{rs}
&=
\frac{\widehat{\mathrm{Cov}}_{\mathcal{P}}
(V_r^{(1)},V_s^{(1)})}{m}
\\
&\quad+
\frac{\widehat{\mathrm{Cov}}_{\mathcal{N}}
(V_r^{(0)},V_s^{(0)})}{n_0},
\\
&\hspace{3em} r,s\in\{a,b\}.
\end{aligned}
\]
With contrast vector $c=(1,-1)$, the estimated variance of the AUC difference
is $\widehat{\mathrm{Var}}(\widehat{\Delta})=c^\top\widehat{\Sigma}c$. We then
form the Wald statistic
\[
Z =
\frac{\widehat{\Delta}}
{\sqrt{\widehat{\mathrm{Var}}(\widehat{\Delta})}}.
\]
Under $H_0$ and standard regularity conditions for U-statistics, $Z$ is
asymptotically standard normal, yielding the two-sided $p$-value
\[
p_{\mathrm{DeLong}}=2\{1-\Phi(|Z|)\}.
\]
This is the primary hypothesis test for ROC-AUC differences in the controlled
experiment table.

\paragraph{Paired bootstrap tests for PR-AUC, F1 score, and recall.}
For PR-AUC, F1 score, and recall, we use a paired nonparametric bootstrap at
the patient/example level because these metrics are nonsmooth or nonlinear
functionals of the empirical distribution \citep{efron1993introduction}. For bootstrap replicate
$b=1,\ldots,B$, draw indices $I_1^{(b)},\ldots,I_n^{(b)}$ independently from
the empirical distribution on $\{1,\ldots,n\}$, reuse the same resampled
indices for all experimental conditions, and compute the paired contrast
$\widehat{\Delta}^{(b)}
=\widehat{\theta}^{(b)}_a-\widehat{\theta}^{(b)}_b$ on the resampled test set.
We report the bootstrap standard error, percentile confidence interval, and a
two-sided sign-based bootstrap $p$-value
\[
p_{\mathrm{boot}}
=2\min\left\{
\begin{aligned}
&
\frac{1+\sum_{b=1}^{B}\mathbb{I}(\widehat{\Delta}^{(b)}\le 0)}{B+1},
\\
&
\frac{1+\sum_{b=1}^{B}\mathbb{I}(\widehat{\Delta}^{(b)}\ge 0)}{B+1}
\end{aligned}
\right\},
\]
truncated at one if necessary. This test asks whether the paired resampling
distribution of the metric difference is centered away from zero, while sharing
resampled indices preserves the correlation induced by common test examples.

\paragraph{Multiple comparisons.}
The full experiment grid contains many controlled comparisons. When making
claims across multiple factors, we interpret isolated small $p$-values
cautiously and report effect sizes together with uncertainty estimates. If a
single family of confirmatory comparisons is specified, the corresponding
$p$-values can be adjusted using Holm--Bonferroni control of the family-wise
error rate \citep{holm1979simple} or Benjamini--Hochberg control of the false
discovery rate \citep{benjamini1995controlling}. In all
cases, statistical significance is interpreted together with the magnitude and
uncertainty of the estimated effect.

\section{Training}
\label{app:training}

\begin{table*}[!t]
\centering
\caption{SFT controlled experiment grid. Rows list factor
levels rather than individual configurations; the SFT sweep crosses all listed data,
training, inference, and metric factors, yielding 504 configurations.}
\label{tab:controlled-experiments}
\scriptsize
\setlength{\tabcolsep}{2pt}
\renewcommand{\arraystretch}{1.15}
\begin{minipage}{\textwidth}
\resizebox{\textwidth}{!}{%
\begin{tabular}{
  >{\centering\arraybackslash}m{0.15\linewidth}
  >{\centering\arraybackslash}m{0.12\linewidth}
  >{\centering\arraybackslash}m{0.12\linewidth}
  >{\centering\arraybackslash}m{0.12\linewidth}
  >{\centering\arraybackslash}m{0.15\linewidth}
  >{\centering\arraybackslash}m{0.30\linewidth}
}
\toprule
\multicolumn{2}{c}{Data} &
\multicolumn{2}{c}{Training} &
\multicolumn{1}{c}{Inference} &
\multicolumn{1}{c}{Metric} \\
\cmidrule(lr){1-2}\cmidrule(lr){3-4}\cmidrule(lr){5-5}\cmidrule(lr){6-6}
\shortstack{Rationale\\format} & Sample size & Learning rate & Model &
\shortstack{(Sampling method\\$*$ Temperature)} &
\shortstack{(Evaluation metric, Hypothesis test)} \\
\midrule
\shortstack{Label\\Only} &
15,323 ($\approx$15.3K) &
$5{\times}10^{-5}$ &
\shortstack{Qwen2.5-7B\\-Instruct} &
Greedy &
\shortstack{(ROC-AUC, Paired DeLong test)} \\
\addlinespace
\shortstack{Free\\rationale} &
3,831 ($\approx$3.8K) &
$1.5{\times}10^{-4}$ &
Qwen3-8B &
(Top-$k$ $*$ 0.1) &
\shortstack{(PR-AUC, Paired bootstrap test)} \\
\addlinespace
\shortstack{Stepwise\\rationale} &
1,533 ($\approx$1.5K) &
$2.5{\times}10^{-4}$ &
 &
(Top-$k$ $*$ 0.5) &
\shortstack{(F1 score, Paired bootstrap test)} \\
\addlinespace
 &
 &
$3.5{\times}10^{-4}$ &
 &
(Top-$k$ $*$ 1.0) &
\shortstack{(Recall, Paired bootstrap test)} \\
\addlinespace
 &
 &
 &
 &
(Top-$p$ $*$ 0.1) &
 \\
\addlinespace
 &
 &
 &
 &
(Top-$p$ $*$ 0.5) &
 \\
\addlinespace
 &
 &
 &
 &
(Top-$p$ $*$ 1.0) &
 \\
\bottomrule
\end{tabular}
}
\end{minipage}
\end{table*}

\subsection{Additional Details}

All experiments in the main SFT sweep used the same longitudinal ADRD
prediction task and the same input representation. Each example contained the
individual's sex and a \texttt{medical history} object,
where each non-missing event was represented as an \texttt{event: age} pair.
The output field, \texttt{ADRD}, stored the training target. We compared three
rationale formats. In Label Only, the target contained
an empty \texttt{<THINK>} block followed by the binary label. In free-rationale, the
empty block was replaced by a concise biomedical rationale. In
stepwise-rationale, the generated rationale was constrained to three steps:
identifying relevant events, aggregating risk, and concluding whether the record
supported ADRD within five years.

\paragraph{SFT prompt.}
The chat prompt used for SFT and for SFT evaluation was shared across all
rationale conditions. The user message inserted the structured record after the
first line shown below. The assistant completion was taken directly from the
\texttt{ADRD} field of the corresponding training record.

\begin{lstlisting}[basicstyle=\ttfamily\scriptsize,breaklines=true,breakatwhitespace=false,columns=fullflexible]
System:
This will not be used to make decisions about a patient. This is for research purposes only.
You are a healthcare risk assessment assistant.
You will be given a single individual in JSON format containing {{event: age}} pairs. Events may include sex, laboratory results, and past diseases/diagnoses encoded with ICD-10 codes.
Your task is to predict whether the individual will develop Alzheimer's disease or related dementias (ADRD) within five years after the last recorded age.
For this task, ADRD is defined by the following ICD-10 codes: F00 (Dementia in Alzheimer's disease), F01 (Vascular dementia), F03 (Unspecified dementia), G30 (Alzheimer's disease), and G31 (Other degenerative diseases of the nervous system, not elsewhere classified).

You MUST follow the output format exactly and output nothing else.

OUTPUT FORMAT:
<THINK>
{THINKING_STEPS}
</THINK>

Prediction (0 or 1): {FINAL_LABEL}

Where:
- {THINKING_STEPS} contains your step-by-step reasoning.
- {FINAL_LABEL} is exactly one character: 0 or 1, where 1 indicates ADRD within 5 years and 0 indicates no ADRD within 5 years.

User:
Here is the input for the individual:
{INPUT_JSON}

Return the output in the required format.
\end{lstlisting}

\paragraph{Rationale generation.}
The rationales used in the free-rationale and stepwise-rationale conditions were
generated from the original training labels before SFT. The
reasoning generator was given the structured patient record, the last observed
age, and the ground-truth \texttt{ADRD\_within\_5y} label. It was instructed to
use only evidence present in the record and not to invent symptoms, imaging,
medications, family history, genetic findings, or lifestyle factors. Long
records were capped at 200 events when constructing the reasoning-generation
prompt. The default reasoning-generation model was GPT-5.4 with medium
reasoning effort and a 450-token output limit. The generated text was then
wrapped as
\texttt{<THINK> reasoning </THINK>} followed by \texttt{Prediction (0 or 1):
label}.

For the free-rationale condition, the generator prompt was:

\begin{lstlisting}[basicstyle=\ttfamily\scriptsize,breaklines=true,breakatwhitespace=false,columns=fullflexible]
You are a clinical epidemiology assistant helping generate training rationales.
You will be given a single individual in JSON format containing {{event: age}} pairs. Events may include sex, laboratory results, and past diseases/diagnoses encoded with ICD-10 codes. Your task is to write a medically plausible reasoning narrative explaining why the ground-truth label ADRD_within_5y is {label} for this individual, based ONLY on the provided events.

Definition: ADRD is defined by the following ICD-10 codes:
- F00 (Dementia in Alzheimer's disease)
- F01 (Vascular dementia)
- F03 (Unspecified dementia)
- G30 (Alzheimer's disease)
- G31 (Other degenerative diseases of the nervous system, not elsewhere classified)

Inputs:
- Patient record: JSON {{event: age}} pairs
- Last observed age: {last_age}
- Ground-truth ADRD_within_5y label: {label}

Instructions:
- Use ONLY evidence present in the record; do NOT invent symptoms, imaging, medications, family history, genetics, or lifestyle.
- You may interpret an ICD-10 code into a condition ONLY if you are highly confident; otherwise keep it as the code and reason generally.
- Focus on biomedical plausibility: age effects; vascular/metabolic risks (e.g., hypertension/diabetes/dyslipidemia); cerebrovascular disease; chronic kidney disease; depression; traumatic brain injury; sleep disorders; and other comorbidities if present in the record.
- If the label is 1, explain how the observed pattern could plausibly precede ADRD within about 5 years after the last age.
- If the label is 0, explain why the record lacks strong indicators of near-term ADRD, or why risk appears lower, while acknowledging uncertainty.
- Keep it coherent and concise: 6-8 sentences, one paragraph. Avoid lists unless necessary.

Output format constraints:
- Output ONLY the reasoning narrative text.
- Do NOT output the label.
- Do NOT output "Prediction".
- Do NOT include <THINK> tags.
- Do NOT add any headings or extra metadata.

Patient record (JSON):
{patient_json}
\end{lstlisting}

For the stepwise-rationale condition, the generator prompt used the same task
definition, but replaced the open narrative instruction with the following
three-step structure:

\begin{lstlisting}[basicstyle=\ttfamily\scriptsize,breaklines=true,breakatwhitespace=false,columns=fullflexible]
Subtasks (write them as Step 1/Step 2/Step 3):
Step 1: Identify relevant events in the record that correlate with ADRD risk using ONLY what appears. If you are not confident interpreting an ICD-10 code, keep it as the code and reason generally.
Step 2: Aggregate risk: weigh age, sex (if present), and the identified events. Explain why they increase/decrease near-term ADRD risk. If evidence is weak/ambiguous, explicitly note uncertainty.
Step 3: Conclude whether ADRD is expected within 5 years after the last recorded age, and make the conclusion consistent with label={label}.

Strict rules:
- Use ONLY evidence present in the JSON; do NOT invent symptoms, imaging, medications, family history, genetics, or lifestyle factors.
- Structure your output as exactly three sections labeled "Step 1:", "Step 2:", and "Step 3:".
- Write 1-2 sentences per step, so 3-6 sentences total.
- Keep the reasoning coherent and biomedically plausible, explicitly tying statements to events/ages in the record.
- Output ONLY the three-step thinking text, with no extra headers, no metadata, no label, and no <THINK> tags.
\end{lstlisting}

\paragraph{Fine-tuning and evaluation.}
The SFT sweep used Qwen/Qwen3-8B and Qwen/Qwen2.5-7B-Instruct, the three
rationale formats above, a cosine learning-rate schedule, and learning
rates of $5\times10^{-5}$, $1.5\times10^{-4}$, $2.5\times10^{-4}$, and
$3.5\times10^{-4}$. Training used QLoRA with 4-bit NF4 quantization, LoRA rank
16, LoRA alpha 32, dropout 0.05, target modules
\texttt{q\_proj}, \texttt{k\_proj}, \texttt{v\_proj}, \texttt{o\_proj},
\texttt{up\_proj}, \texttt{down\_proj}, and \texttt{gate\_proj}, batch size
one, gradient accumulation over eight steps, three epochs, warmup ratio 0.03,
maximum sequence length 2048, and completion-only loss. For SFT evaluation, we
used the same held-out test records for all rationale conditions.
The score for ROC-AUC and PR-AUC was the sigmoid of the generated-label logit
difference, $\sigma(\ell_1-\ell_0)$, at the generation step where the final
\texttt{0} or \texttt{1} label was produced.

\paragraph{CVD validation protocol.}
The cross-disease check defines CVD by the first recorded occurrence of
ICD-10 codes I20--I25 or I60--I64, uses a five-year prediction window with a one-year offset, and
excludes those target codes from the input. Label Only and Rationale First
free-rationale Qwen3-8B SFT use identical subsets of 1,533 training records and
853 test records (171 cases and 682 controls), learning rate
$1.5\times10^{-4}$, three epochs, seed 42, and greedy decoding. The free
rationales are generated with access to the CVD label and use the same
evidence-only constraints as the ADRD rationales. ROC-AUC and PR-AUC use the
generated-label logit score. The
Label Only/free-rationale results are 0.669/0.632 ROC-AUC, 0.344/0.302 PR-AUC,
0.056/0.181 F1, and 0.029/0.123 recall, with no unparsed outputs.

\subsection{Additional Results}

\begin{figure*}[!p]
\centering
\captionsetup[subfigure]{skip=1pt}
\begin{subfigure}[t]{0.31\textwidth}
\centering
\includegraphics[width=\linewidth]{figures/sft_roc-auc_by_thinking_v0.pdf}
\caption{Rationale, ROC-AUC}
\end{subfigure}
\hspace{5mm}
\begin{subfigure}[t]{0.48\textwidth}
\centering
\includegraphics[width=\linewidth]{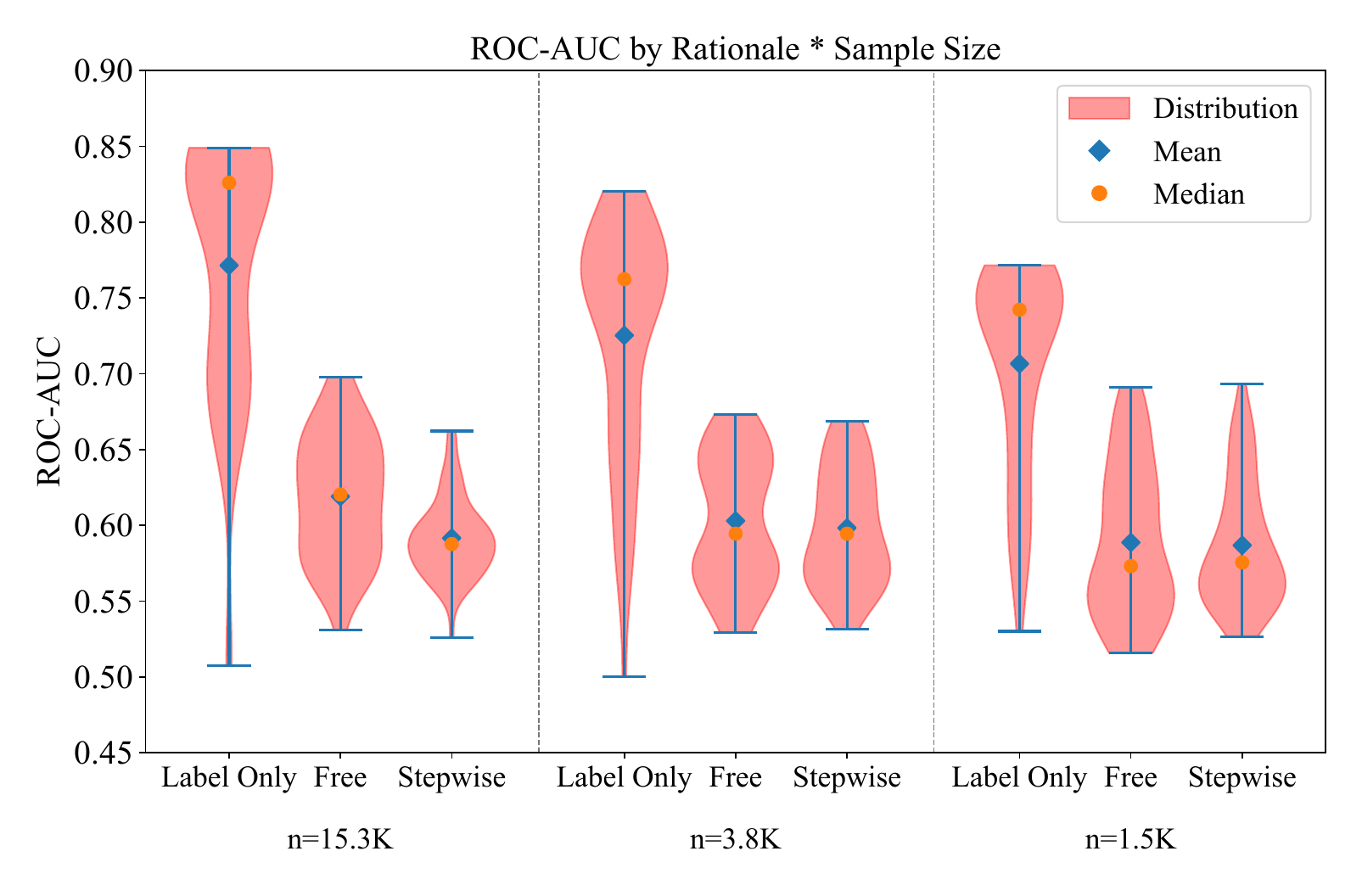}
\caption{Rationale and sample size, ROC-AUC}
\end{subfigure}

\vspace{-0.25em}

\begin{subfigure}[t]{0.31\textwidth}
\centering
\includegraphics[width=\linewidth]{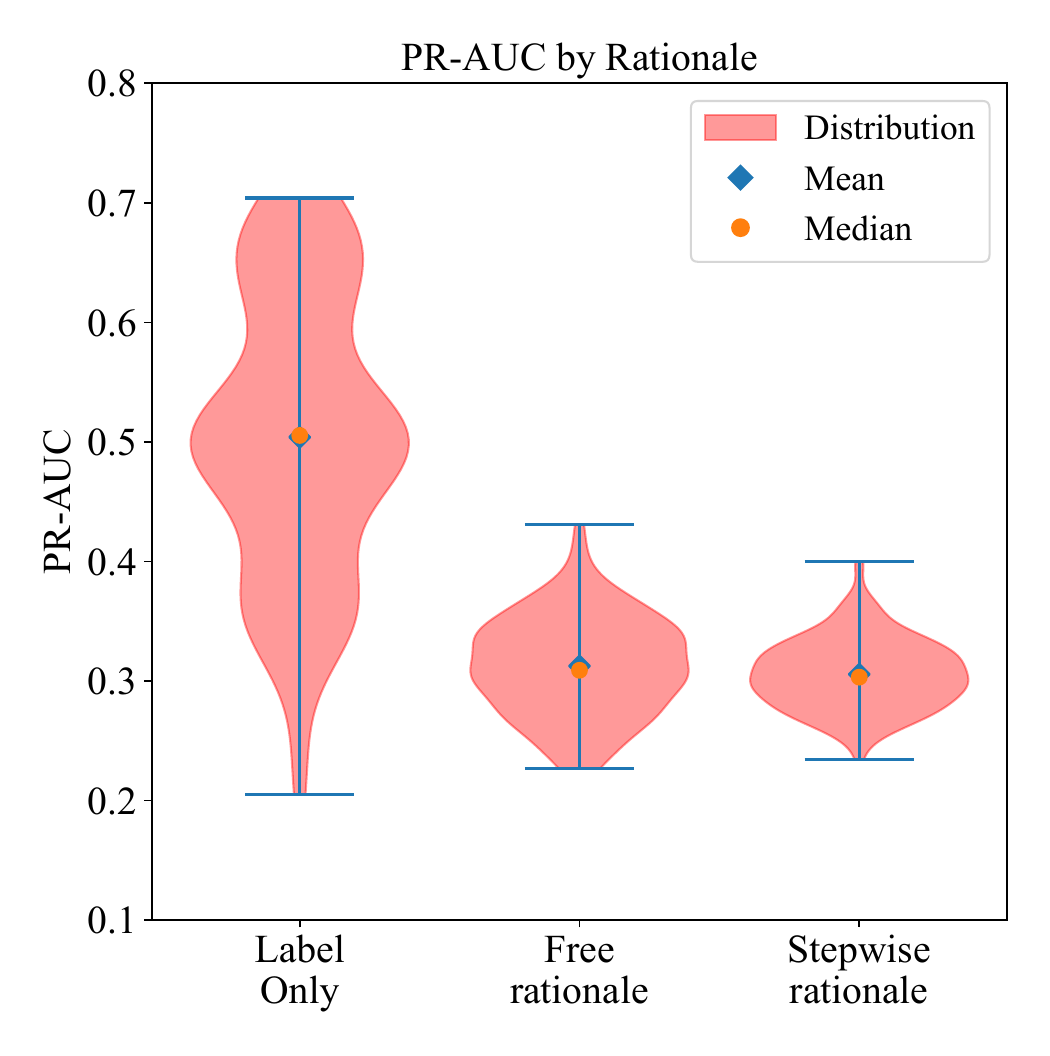}
\caption{Rationale, PR-AUC}
\end{subfigure}
\hspace{5mm}
\begin{subfigure}[t]{0.48\textwidth}
\centering
\includegraphics[width=\linewidth]{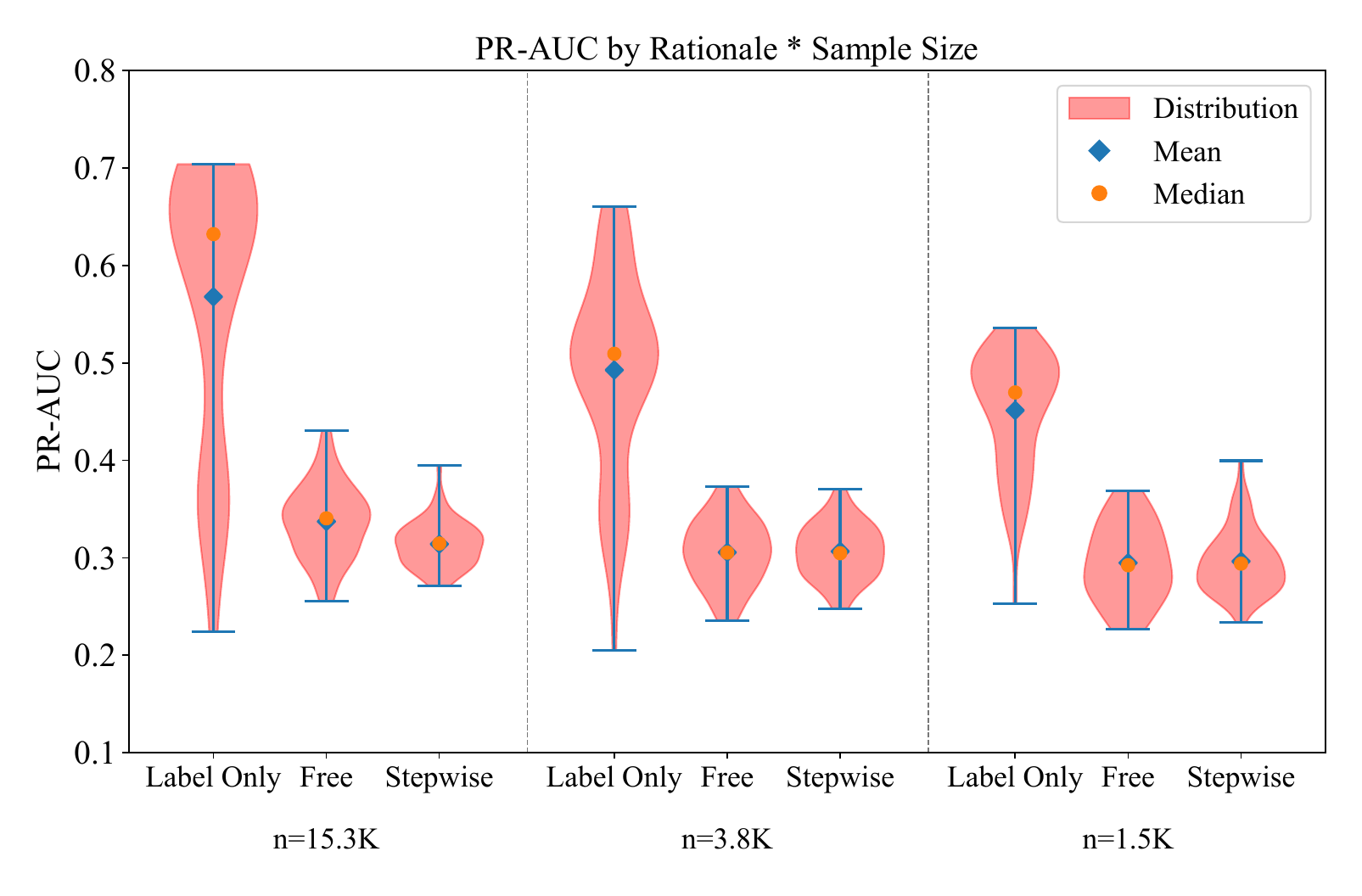}
\caption{Rationale and sample size, PR-AUC}
\end{subfigure}

\vspace{-0.25em}

\begin{subfigure}[t]{0.31\textwidth}
\centering
\includegraphics[width=\linewidth]{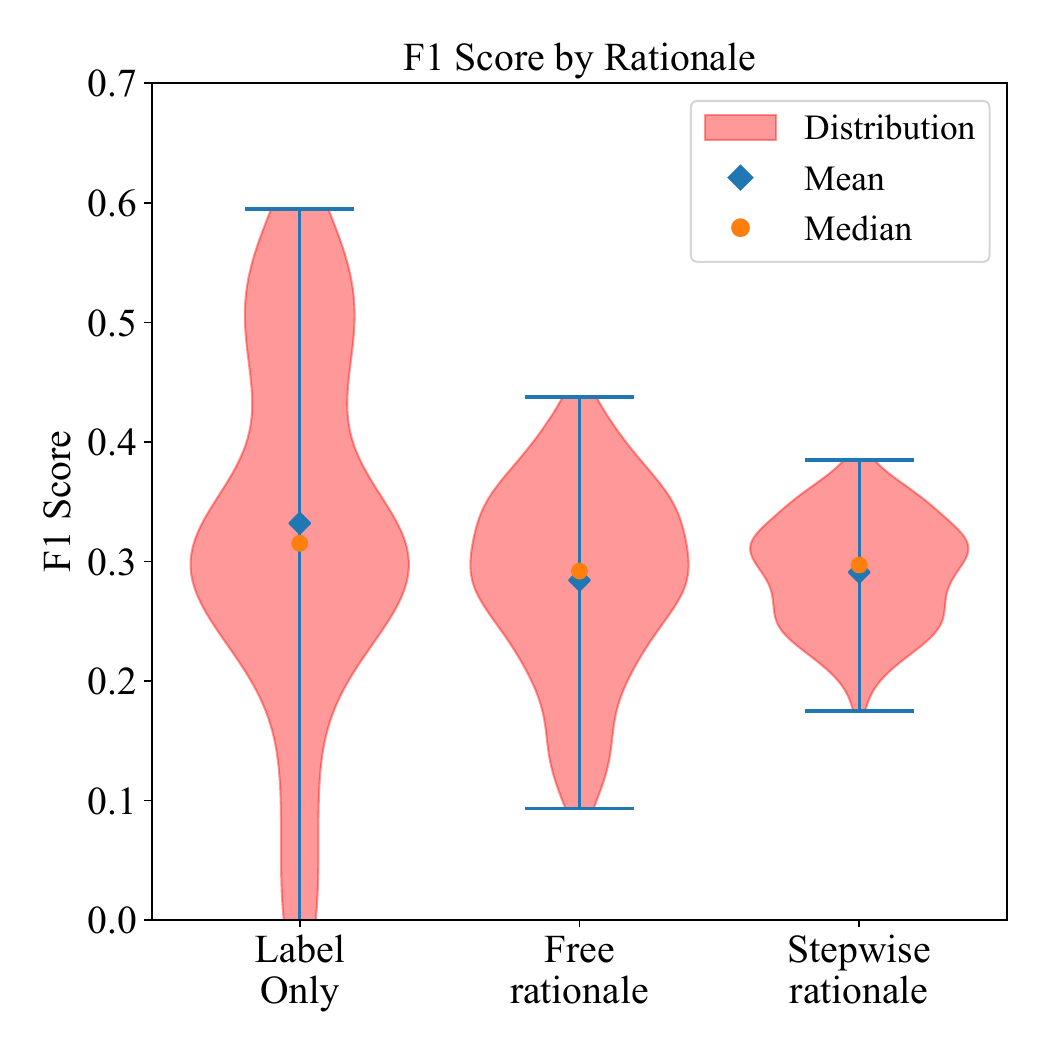}
\caption{Rationale, F1 score}
\end{subfigure}
\hspace{5mm}
\begin{subfigure}[t]{0.48\textwidth}
\centering
\includegraphics[width=\linewidth]{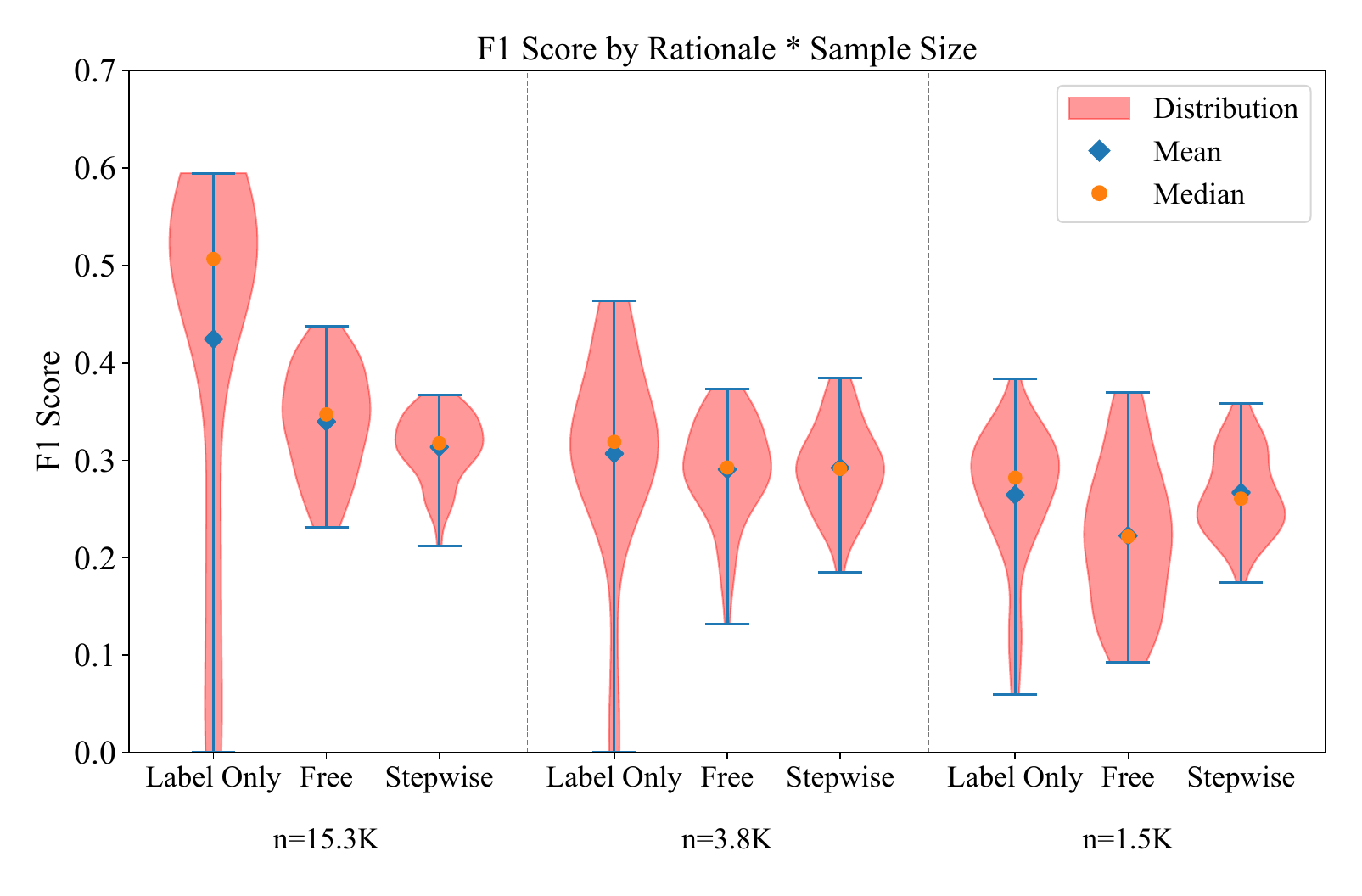}
\caption{Rationale and sample size, F1 score}
\end{subfigure}

\vspace{-0.25em}

\begin{subfigure}[t]{0.31\textwidth}
\centering
\includegraphics[width=\linewidth]{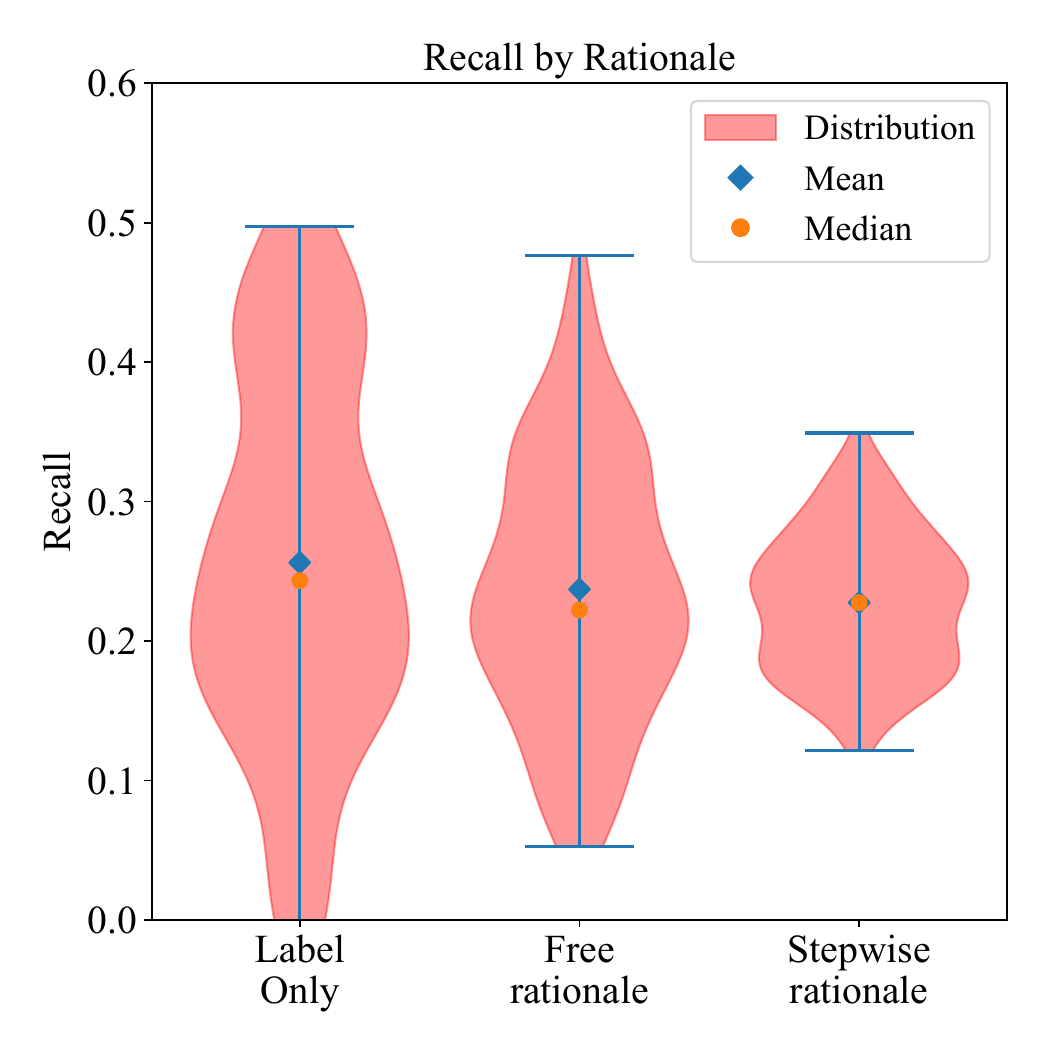}
\caption{Rationale, Recall}
\end{subfigure}
\hspace{5mm}
\begin{subfigure}[t]{0.48\textwidth}
\centering
\includegraphics[width=\linewidth]{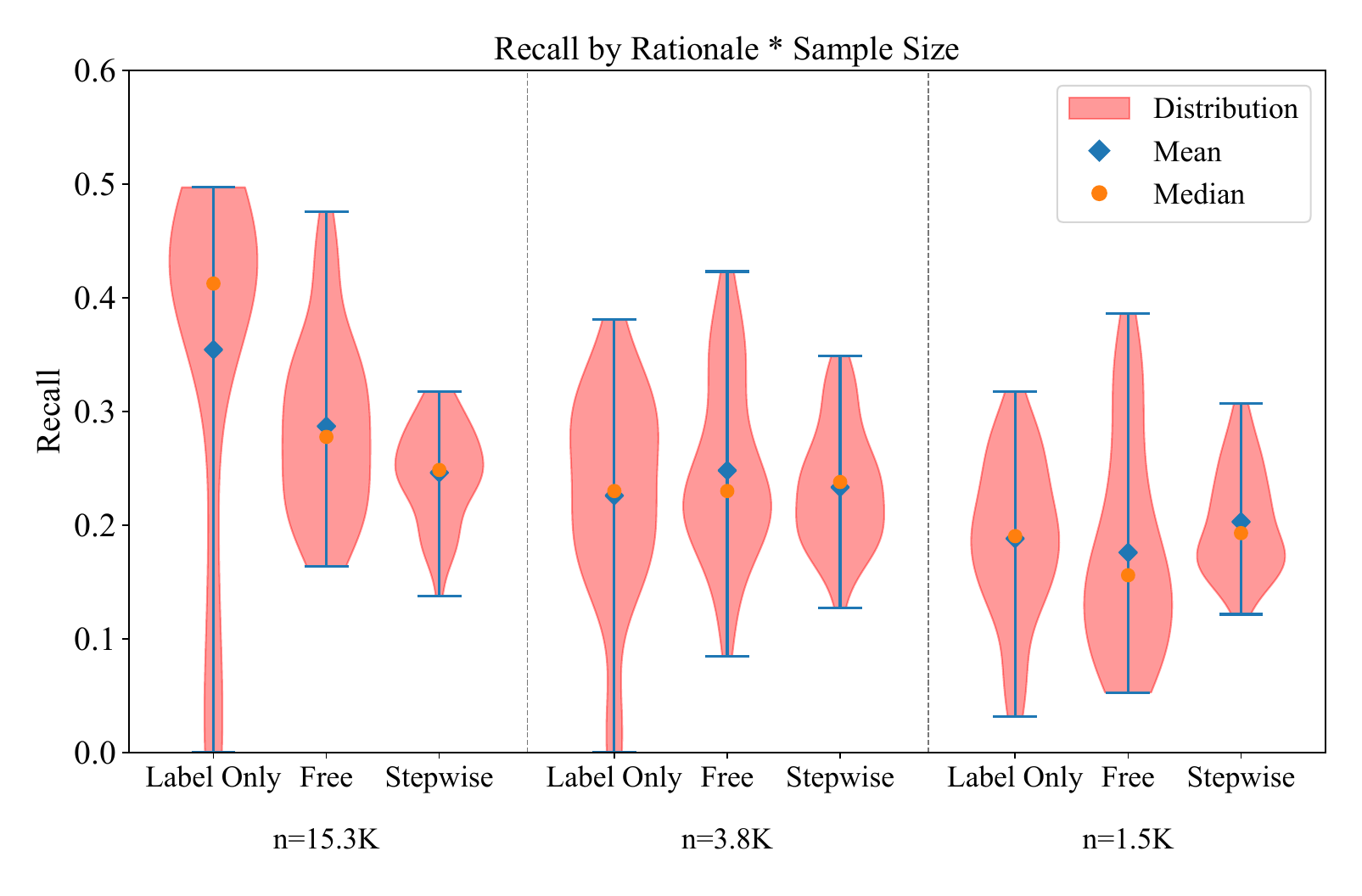}
\caption{Rationale and sample size, Recall}
\end{subfigure}
\caption{SFT performance by rationale format and by rationale format crossed
with training sample size. Panel A repeats the rationale-format ROC-AUC summary
from Figure~\ref{fig:sft-rationale-ranking}A for comparison with the appendix
sample-size and non-ROC panels.}
\label{fig:app-sft-rationale-roc-samplesize}
\label{fig:app-sft-rationale-pr}
\label{fig:app-sft-rationale-thresholded}
\label{fig:app-sft-thinking-f1-recall}
\label{fig:app-sft-thinking-samplesize-f1-recall}
\end{figure*}

For rationale comparisons, the non-ROC metrics followed the same overall direction as the main ROC-AUC result. Label Only had higher mean PR-AUC than free-rationale (0.504 versus 0.313; paired $t$-test, $P=1.79\times10^{-49}$) and stepwise-rationale (0.504 versus 0.306; $P=1.92\times10^{-52}$). Label Only also had higher mean F1 score than free- and stepwise-rationale (0.332 versus 0.284 and 0.291; $P=1.03\times10^{-4}$ and $P=4.18\times10^{-4}$). Mean recall was higher for Label Only than for stepwise-rationale (0.256 versus 0.228; $P=3.86\times10^{-3}$), while the recall difference between Label Only and free-rationale was smaller and not statistically significant (0.256 versus 0.237; $P=0.068$).
Figure~\ref{fig:app-sft-rationale-roc-samplesize} summarizes these rationale-format comparisons across ROC-AUC, PR-AUC, F1 score, and recall, with the right-column panels showing the same contrasts within each training sample size.

Among the best SFT configurations selected by ROC-AUC, the Label Only configuration also had better non-ROC metrics than the best free- and stepwise-rationale configurations. In the Label Only versus free-rationale comparison, PR-AUC was 0.701 versus 0.380 ($P=9.99\times10^{-4}$), F1 score was 0.602 versus 0.436 ($P=9.99\times10^{-4}$), and recall was 0.489 versus 0.408 ($P=0.085$). In the Label Only versus stepwise-rationale comparison, PR-AUC was 0.708 versus 0.361, F1 score was 0.607 versus 0.353, and recall was 0.495 versus 0.293; all three differences favored Label Only ($P=9.99\times10^{-4}$ for PR-AUC, F1 score, and recall).
Figure~\ref{fig:app-free-rationale-error-examples} gives two test examples from the Label Only and free-rationale SFT comparison, illustrating how generated rationales can overweight nonspecific risk in a control or underweight distributed risk in an ADRD case.

\begin{figure}[!htbp]
\centering
\scriptsize
\setlength{\tabcolsep}{3pt}
\renewcommand{\arraystretch}{1.08}
\begin{tabularx}{\columnwidth}{@{}>{\raggedright\arraybackslash}p{0.31\columnwidth}X@{}}
\toprule
Case and setting & Evidence or error pattern \\
\midrule
Control, Label Only correct &
E78 lipidaemia, high BMI, smoking, low cognitive/digit bins, and no dementia or major neurovascular code. \\
\addlinespace[0.2em]
Control, free-rationale wrong &
Treats nonspecific lifestyle and weak cognitive risk as sufficient ADRD evidence. \\
\addlinespace[0.2em]
ADRD case, Label Only correct &
I10 hypertension, E11/E14 diabetes, depression, B12 anaemia, high BMI, and very low matching time. \\
\addlinespace[0.2em]
ADRD case, free-rationale wrong &
Discounts vascular, metabolic, psychiatric, anaemia, BMI, and cognitive-timing signals because no explicit dementia ICD code is observed. \\
\bottomrule
\end{tabularx}
\caption{Divergent test examples from the best Label Only and
free-rationale SFT configurations. Label Only SFT is correct in both cases,
whereas free-rationale SFT overweights nonspecific risk in a control and
underweights distributed risk in an ADRD case.}
\label{fig:app-free-rationale-error-examples}
\end{figure}

\begin{figure*}[!p]
\centering
\captionsetup[subfigure]{skip=1pt}
\begin{subfigure}[t]{0.31\textwidth}
\centering
\includegraphics[width=\linewidth]{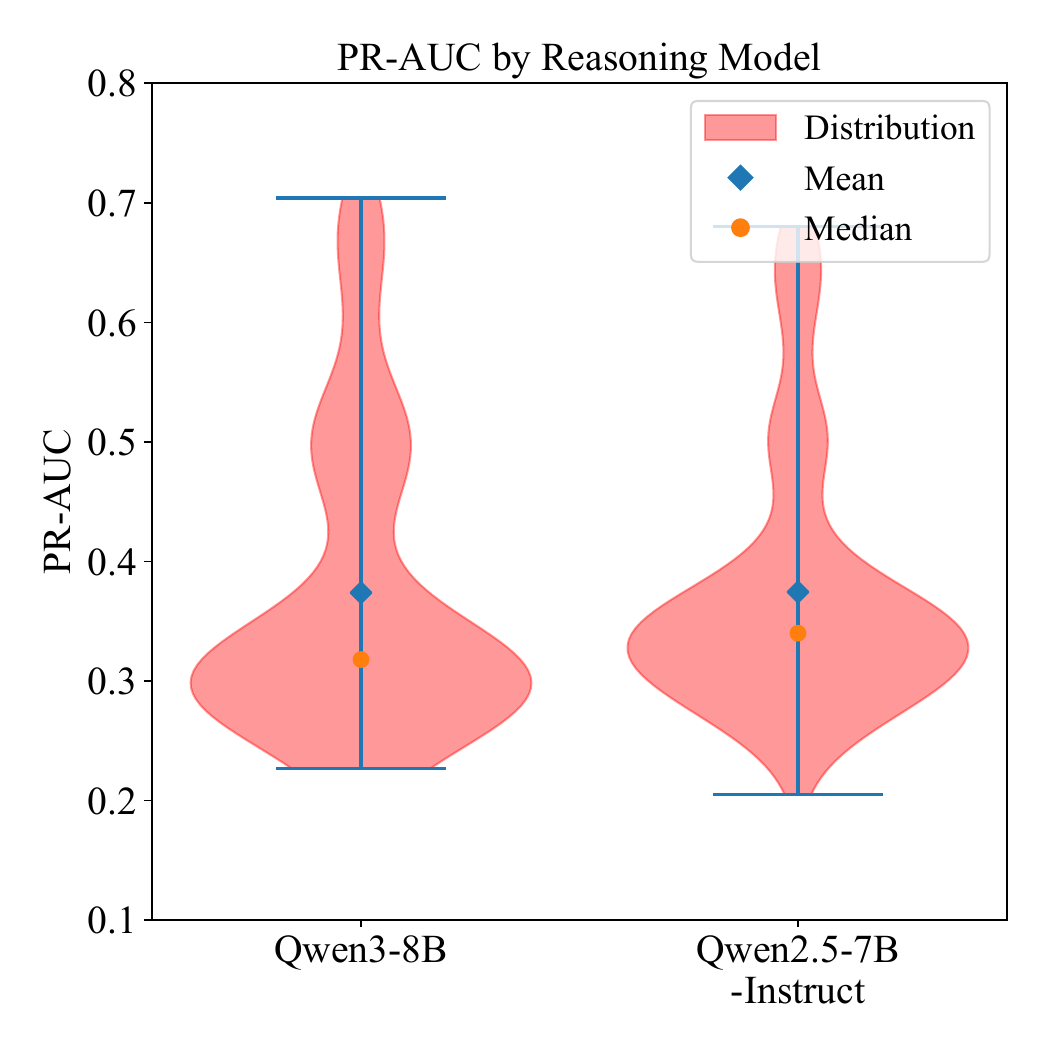}
\caption{Base model, PR-AUC}
\end{subfigure}
\hspace{5mm}
\begin{subfigure}[t]{0.48\textwidth}
\centering
\includegraphics[width=\linewidth]{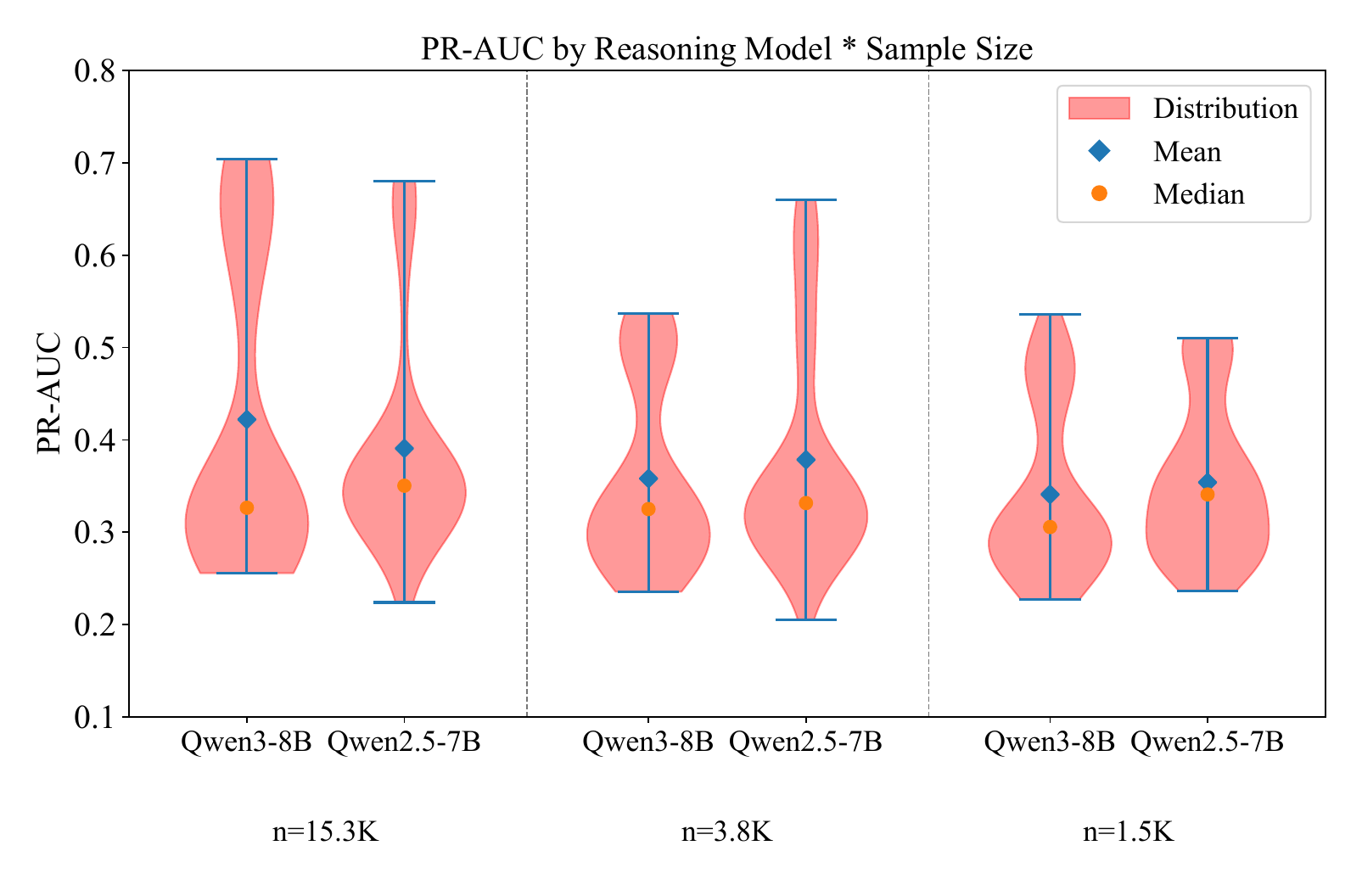}
\caption{Base model and sample size, PR-AUC}
\end{subfigure}

\vspace{-0.1em}

\begin{subfigure}[t]{0.31\textwidth}
\centering
\includegraphics[width=\linewidth]{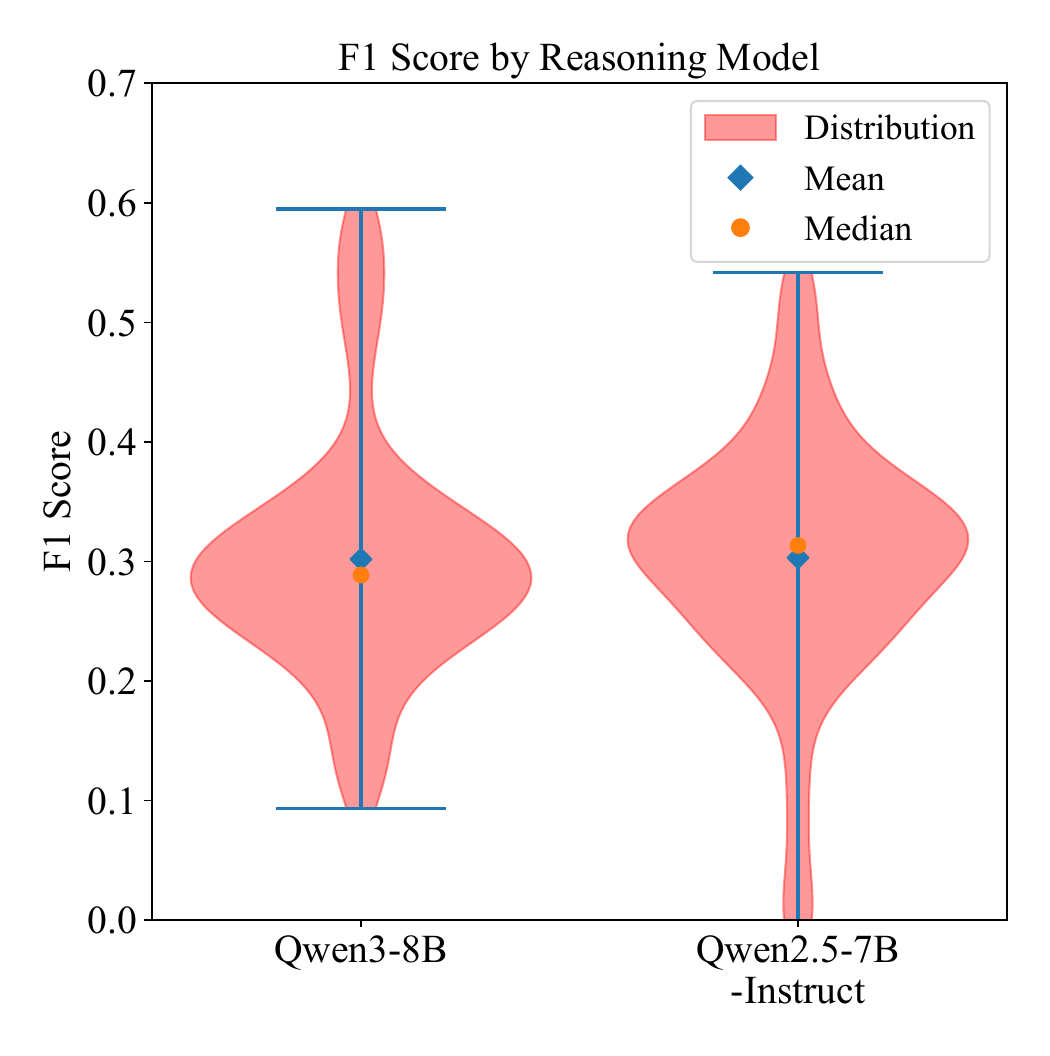}
\caption{Base model, F1 score}
\end{subfigure}
\hspace{5mm}
\begin{subfigure}[t]{0.48\textwidth}
\centering
\includegraphics[width=\linewidth]{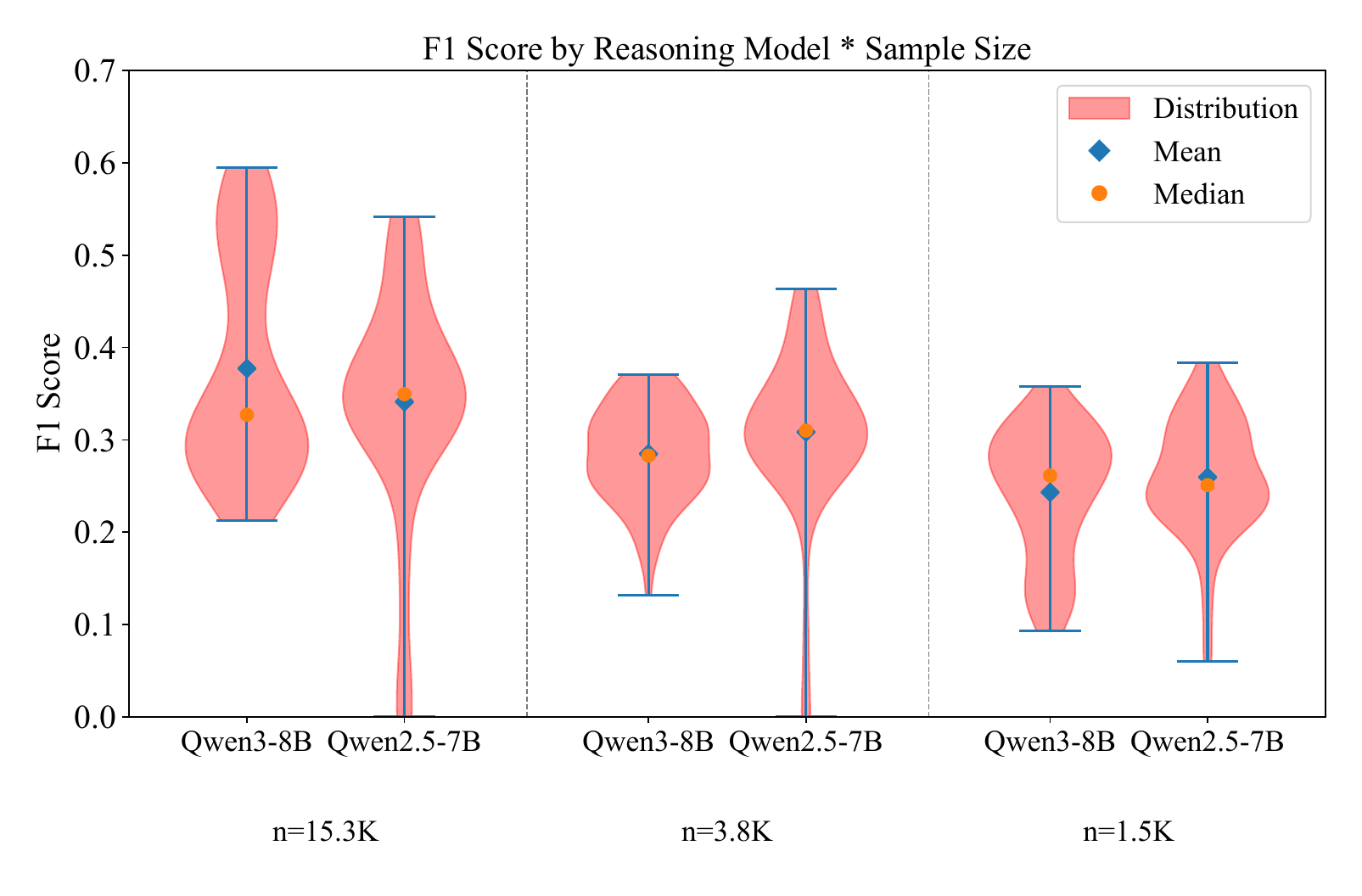}
\caption{Base model and sample size, F1 score}
\end{subfigure}

\vspace{-0.1em}

\begin{subfigure}[t]{0.31\textwidth}
\centering
\includegraphics[width=\linewidth]{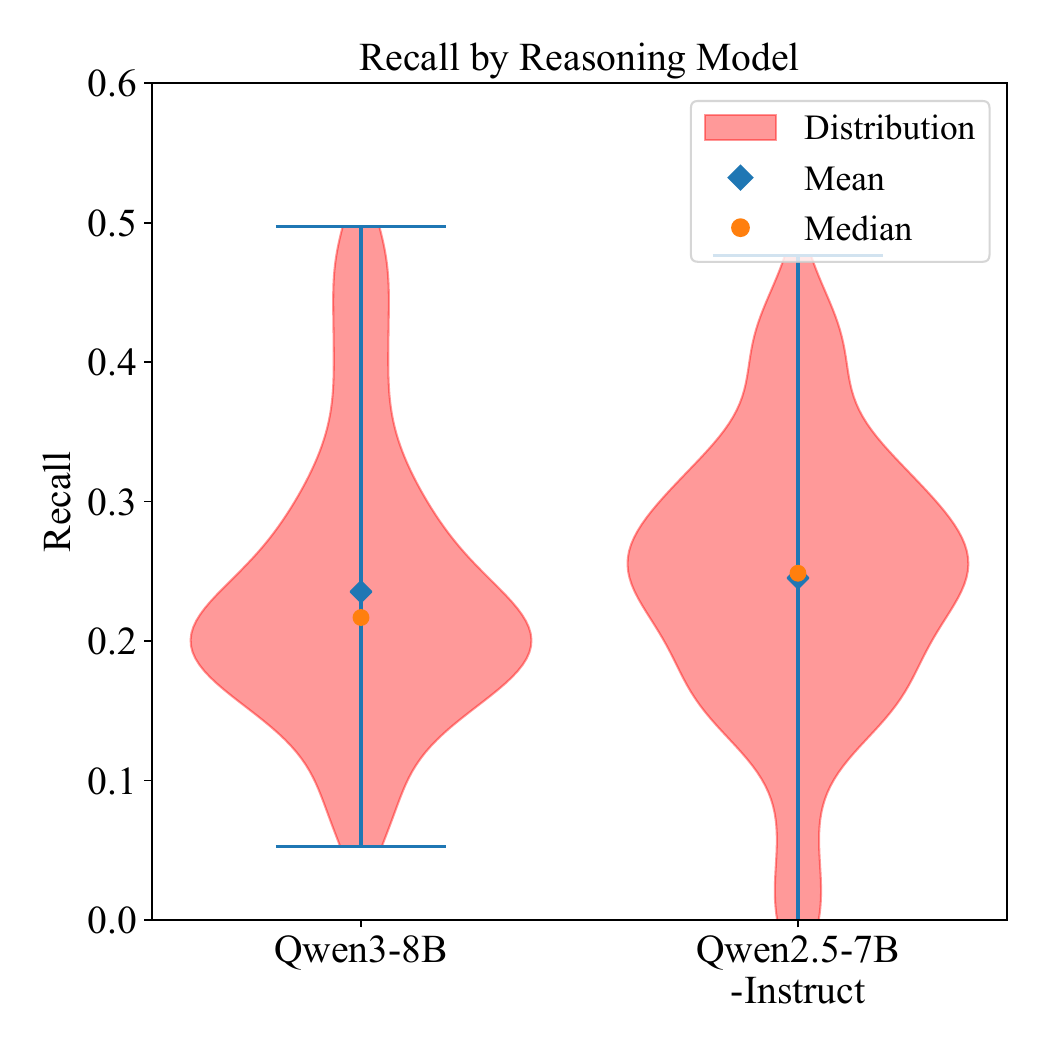}
\caption{Base model, Recall}
\end{subfigure}
\hspace{5mm}
\begin{subfigure}[t]{0.48\textwidth}
\centering
\includegraphics[width=\linewidth]{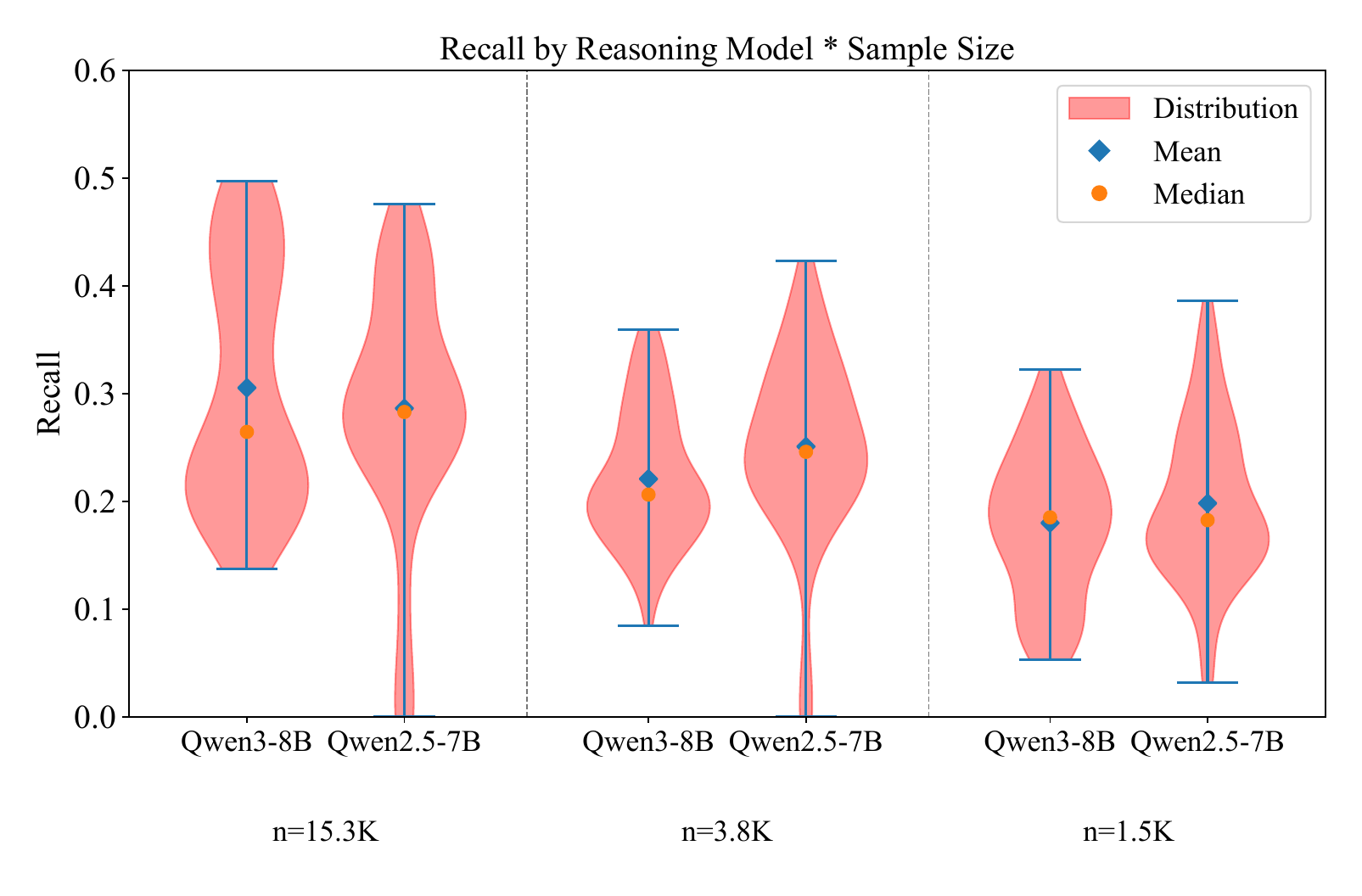}
\caption{Base model and sample size, Recall}
\end{subfigure}
\caption{Additional SFT PR-AUC, F1 score, and Recall by base model and by base model crossed with training sample size. The interaction panels group sample size first and then base model.}
\label{fig:app-sft-model-pr}
\label{fig:app-sft-thresholded}
\label{fig:app-sft-model-f1-recall}
\label{fig:app-sft-model-samplesize-f1-recall}
\end{figure*}

The non-ROC metrics likewise showed no meaningful advantage for Qwen3-8B over Qwen2.5-7B-Instruct. Across matched SFT settings, mean PR-AUC was 0.3737 for Qwen3-8B and 0.3744 for Qwen2.5-7B-Instruct ($P=0.916$), mean F1 score was 0.3018 versus 0.3031 ($P=0.874$), and mean recall was 0.235 versus 0.245 ($P=0.152$). Among the best SFT configurations selected by ROC-AUC, Qwen3-8B had PR-AUC 0.701, F1 score 0.594, and recall 0.471, compared with PR-AUC 0.679, F1 score 0.563, and recall 0.437 for Qwen2.5-7B-Instruct; none of these paired bootstrap comparisons was statistically significant ($P=0.125$, $P=0.155$, and $P=0.169$).
Figure~\ref{fig:app-sft-model-pr} shows PR-AUC, F1, and recall distributions by base model, overall and within each training sample size.

\clearpage

\section{Training-Free}
\label{app:training-free}

\begin{table*}[!t]
\centering
\caption{Training-free zero-shot controlled inference grid. Rows list factor
levels rather than unique configurations; the full grid crosses the prompt setting, base
model, and decoding configuration.}
\label{tab:training-free-controlled-experiments}
\scriptsize
\setlength{\tabcolsep}{2pt}
\renewcommand{\arraystretch}{1.15}
\begin{minipage}{\textwidth}
\resizebox{\textwidth}{!}{%
\begin{tabular}{
  >{\centering\arraybackslash}m{0.18\linewidth}
  >{\centering\arraybackslash}m{0.14\linewidth}
  >{\centering\arraybackslash}m{0.20\linewidth}
  >{\centering\arraybackslash}m{0.40\linewidth}
}
\toprule
\multicolumn{1}{c}{Prompting} &
\multicolumn{2}{c}{Inference} &
\multicolumn{1}{c}{Metric} \\
\cmidrule(lr){1-1}\cmidrule(lr){2-3}\cmidrule(lr){4-4}
Prompt setting & Model & \shortstack{(Sampling method,\\Temperature)}
& \mbox{(Evaluation metric, Hypothesis test)} \\
\midrule
Zero-shot &
\shortstack{Qwen2.5-7B-Instruct} &
Greedy &
\mbox{(ROC-AUC, Paired DeLong test)} \\
\addlinespace
\shortstack{Zero-shot\\with CoT} &
Qwen3-8B &
(Top-$k$, 0.1) &
\mbox{(PR-AUC, Paired bootstrap test)} \\
\addlinespace
 &
 &
(Top-$k$, 0.5) &
\mbox{(F1 score, Paired bootstrap test)} \\
\addlinespace
 &
 &
(Top-$k$, 1.0) &
\mbox{(Recall, Paired bootstrap test)} \\
\addlinespace
 &
 &
(Top-$p$, 0.1) &
 \\
\addlinespace
 &
 &
(Top-$p$, 0.5) &
 \\
\addlinespace
 &
 &
(Top-$p$, 1.0) &
 \\
\bottomrule
\end{tabular}
}
\end{minipage}
\end{table*}

\subsection{Additional Details}

Training-free experiments evaluated the same held-out ADRD prediction task
without parameter updates. We evaluated Qwen2.5-7B-Instruct
(\texttt{qwen/qwen-2.5-7b-instruct}) and Qwen3-8B
(\texttt{qwen/qwen3-8b}) through the OpenRouter chat-completion API. This appendix
reports the zero-shot settings. Few-shot evaluations are reported in
Section~\ref{sec:fewshot-ablation} and are excluded from the zero-shot figures
and tests.

\paragraph{Evaluation set and decoding.}
For each evaluation configuration, the test records were shuffled with seed 42, the first
$20\%+1$ records were retained, and SLURM array jobs evaluated non-overlapping
chunks of 15 records. Deterministic configurations used temperature 0. Sampling configurations
crossed two sampling methods, nucleus sampling with \texttt{top\_p=0.9} and
top-$k$ sampling with \texttt{top\_k=50}, with temperatures 0.1, 0.5, and 1.0.
Each API call allowed up to 1024 generated tokens and was retried up to five
times. For Qwen3-8B, the provider-side reasoning flag was enabled in both
prompt conditions; therefore, the zero-shot with CoT condition in the paper
refers to whether the visible output format asked for a \texttt{<THINK>} block,
not to the provider-side flag. The model output was parsed from the numeric
value after \texttt{Probability:}. If that string was absent, the parser also
accepted a bare numeric output or the last non-empty line when it was a number
between 0 and 1.

\paragraph{Zero-shot prompt.}
The zero-shot prompt asked the model to return only a numeric ADRD
probability:

\begin{lstlisting}[basicstyle=\ttfamily\scriptsize,breaklines=true,breakatwhitespace=false,columns=fullflexible]
System:
This will not be used to make decisions about a patient. This is for research purposes only.
You are a healthcare risk assessment assistant.
You will be given a single individual in JSON format containing {{event: age}} pairs. Events may include sex, laboratory results, and past diseases/diagnoses encoded with ICD-10 codes.
Your task is to predict whether the individual will develop Alzheimer's disease or related dementias (ADRD) within five years after the last recorded age.
For this task, ADRD is defined by the following ICD-10 codes: F00 (Dementia in Alzheimer's disease), F01 (Vascular dementia), F03 (Unspecified dementia), G30 (Alzheimer's disease), and G31 (Other degenerative diseases of the nervous system, not elsewhere classified).

You MUST follow the output format exactly and output nothing else.

OUTPUT FORMAT:
Probability: {PROBABILITY}

Where:
- {PROBABILITY} is exactly one number between 0 and 1.

User:
Here is the input for the individual:
{INPUT_JSON}

Return the output in the required format.
\end{lstlisting}

\paragraph{Zero-shot with CoT prompt.}
The zero-shot with CoT prompt used the same system and user content, but
required the model to include step-by-step reasoning before the probability:

\begin{lstlisting}[basicstyle=\ttfamily\scriptsize,breaklines=true,breakatwhitespace=false,columns=fullflexible]
System:
This will not be used to make decisions about a patient. This is for research purposes only.
You are a healthcare risk assessment assistant.
You will be given a single individual in JSON format containing {{event: age}} pairs. Events may include sex, laboratory results, and past diseases/diagnoses encoded with ICD-10 codes.
Your task is to predict whether the individual will develop Alzheimer's disease or related dementias (ADRD) within five years after the last recorded age.
For this task, ADRD is defined by the following ICD-10 codes: F00 (Dementia in Alzheimer's disease), F01 (Vascular dementia), F03 (Unspecified dementia), G30 (Alzheimer's disease), and G31 (Other degenerative diseases of the nervous system, not elsewhere classified).

You MUST follow the output format exactly and output nothing else.

OUTPUT FORMAT:
<THINK>
{THINKING_STEPS}
</THINK>

Probability: {PROBABILITY}

Where:
- {THINKING_STEPS} contains your step-by-step reasoning.
- {PROBABILITY} is exactly one number between 0 and 1.

User:
Here is the input for the individual:
{INPUT_JSON}

Return the output in the required format.
\end{lstlisting}

The parsed probability was used directly as the score for ROC-AUC and PR-AUC.
For thresholded metrics, probabilities at least 0.5 were counted as predicted
ADRD cases.

\subsection{Additional Results}

\begin{figure*}[!p]
\centering
\begin{subfigure}[t]{0.33\textwidth}
\centering
\includegraphics[width=\linewidth]{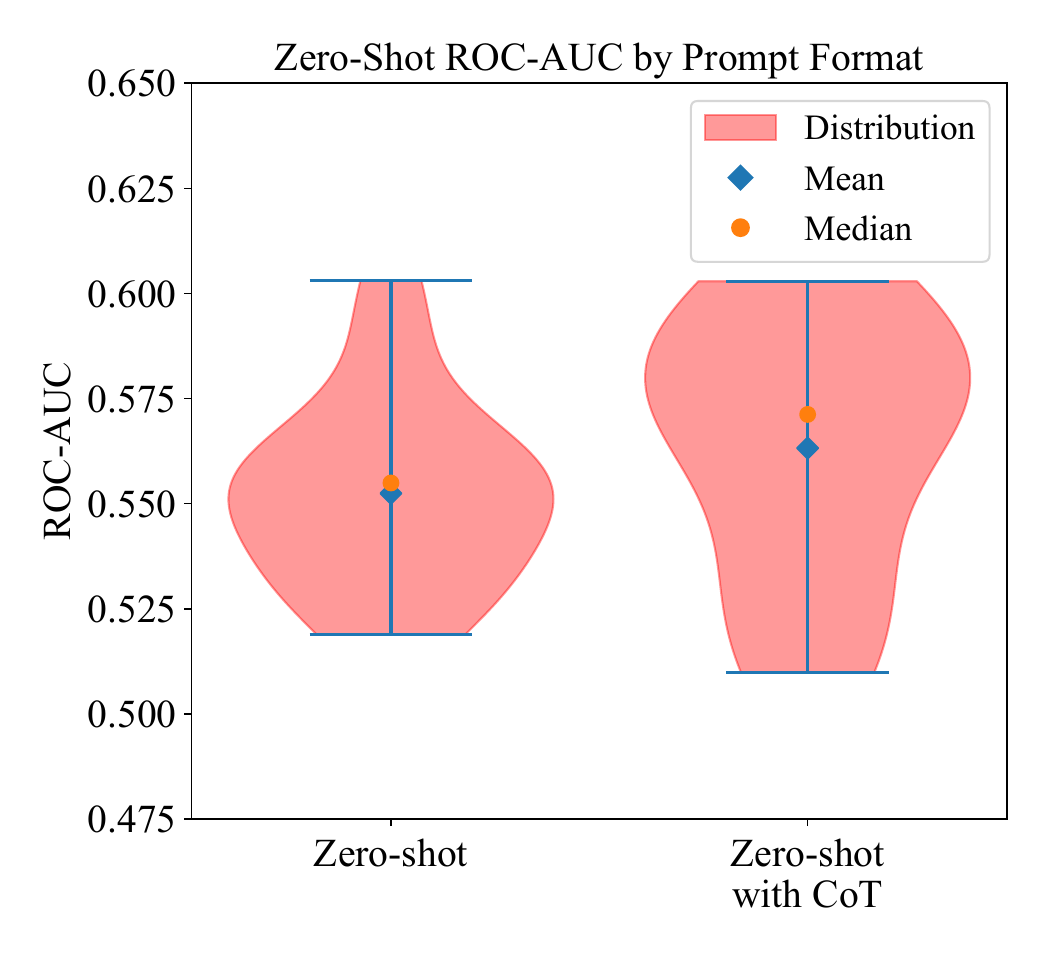}
\caption{Prompt format, ROC-AUC}
\end{subfigure}
\hspace{5mm}
\begin{subfigure}[t]{0.33\textwidth}
\centering
\includegraphics[width=\linewidth]{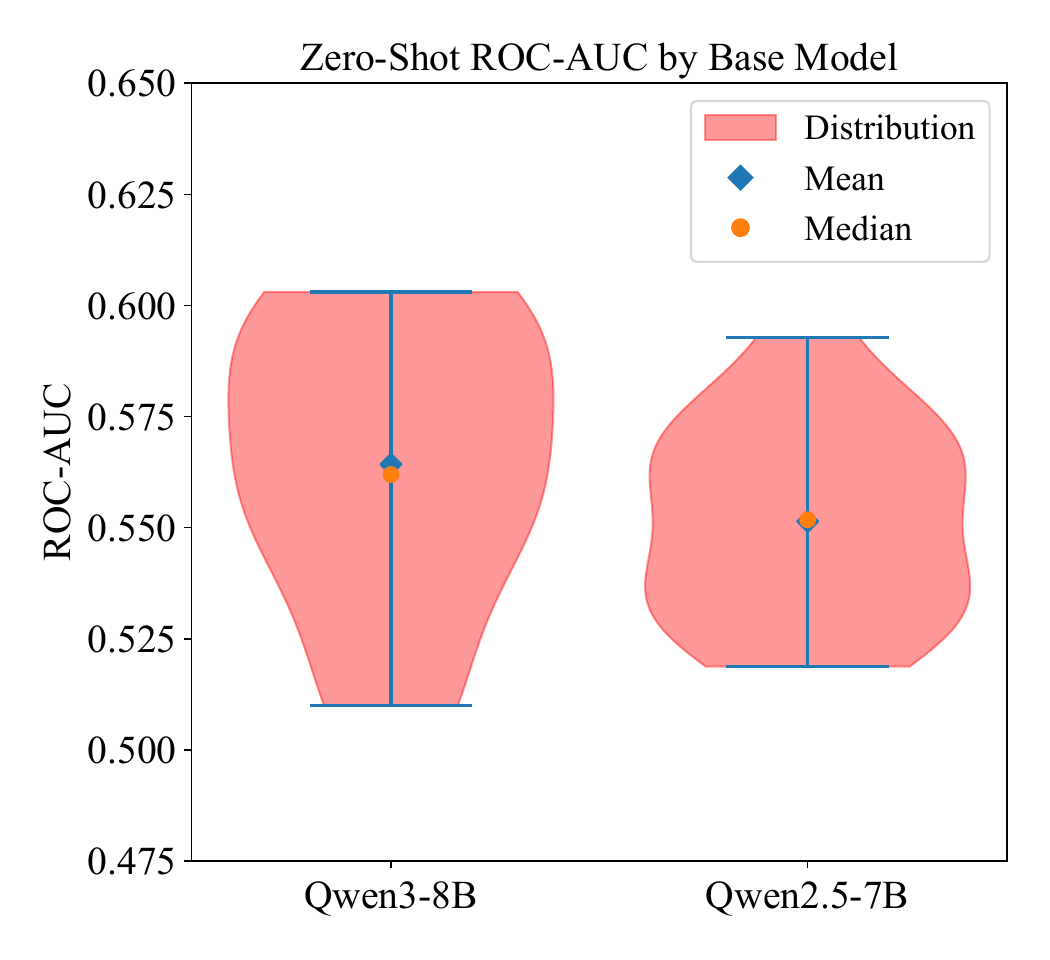}
\caption{Base model, ROC-AUC}
\end{subfigure}

\vspace{0.15em}

\begin{subfigure}[t]{0.33\textwidth}
\centering
\includegraphics[width=\linewidth]{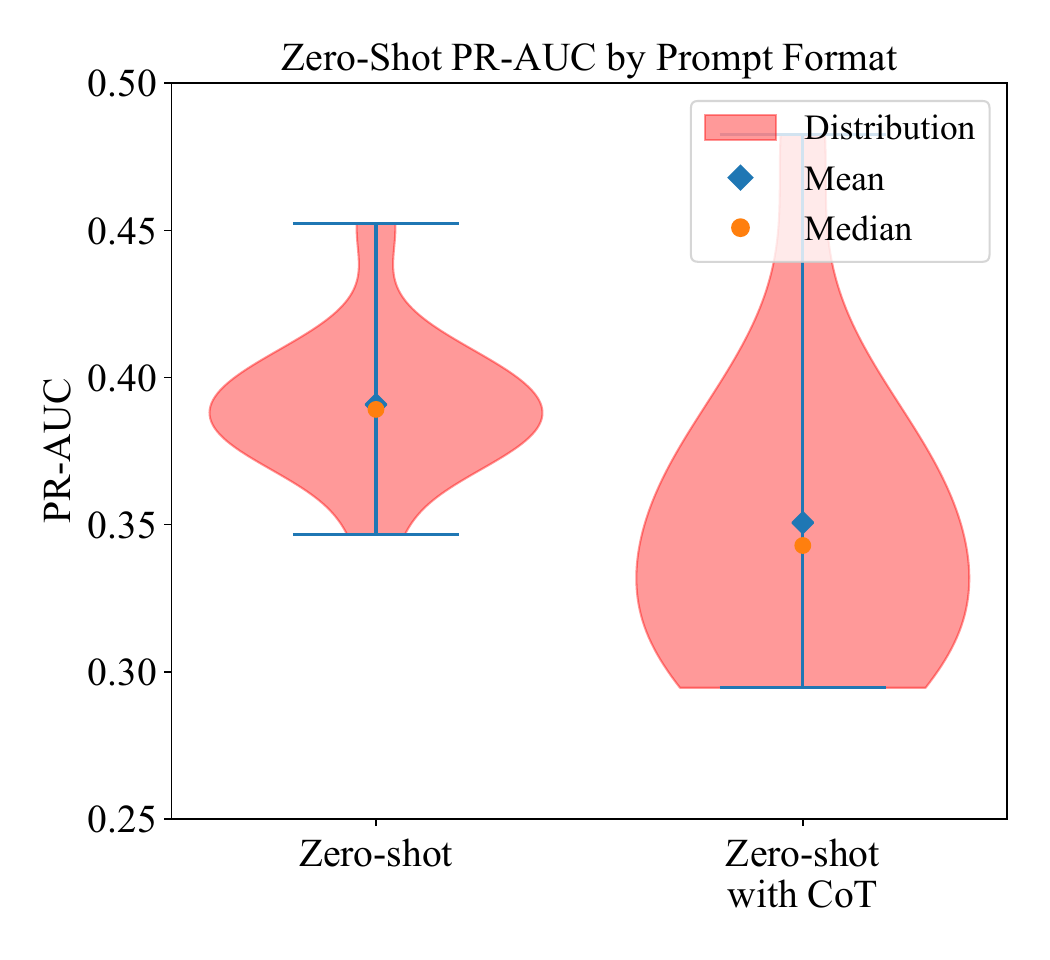}
\caption{Prompt format, PR-AUC}
\end{subfigure}
\hspace{5mm}
\begin{subfigure}[t]{0.33\textwidth}
\centering
\includegraphics[width=\linewidth]{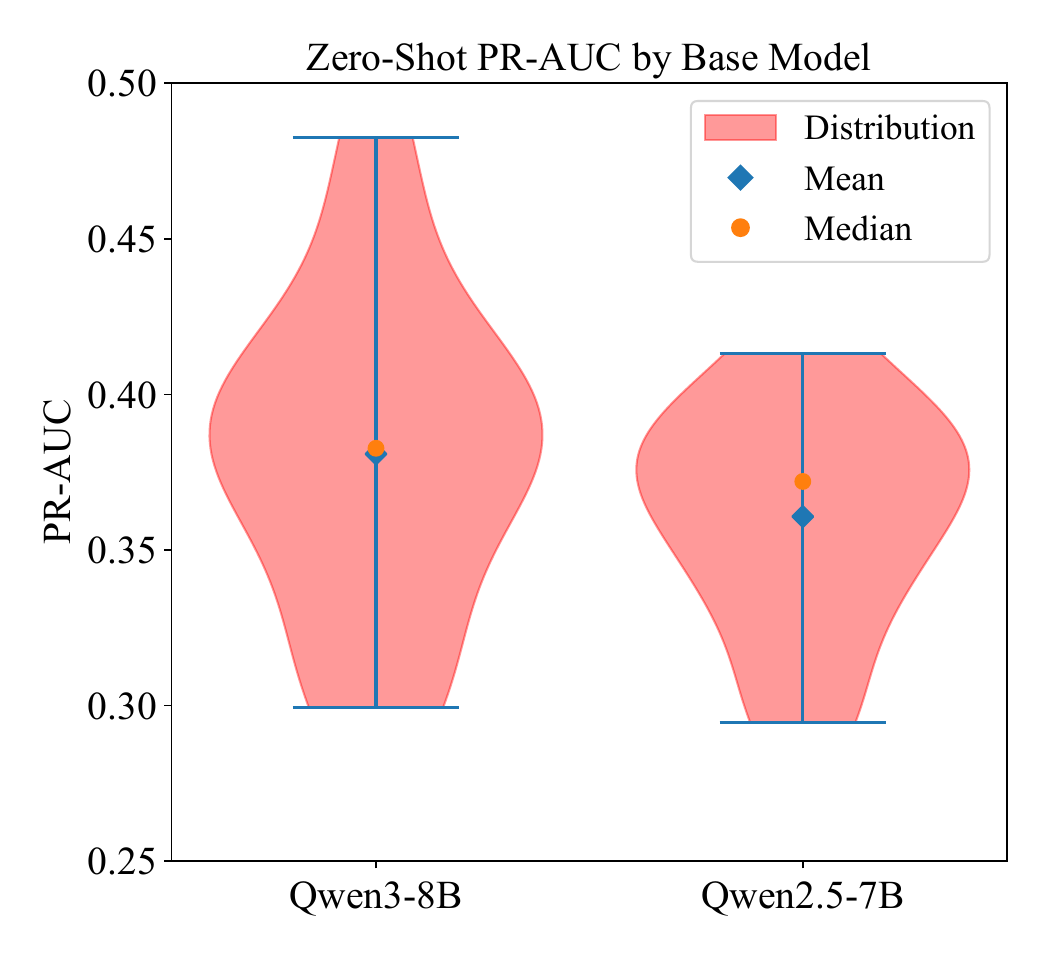}
\caption{Base model, PR-AUC}
\end{subfigure}

\vspace{0.15em}

\begin{subfigure}[t]{0.33\textwidth}
\centering
\includegraphics[width=\linewidth]{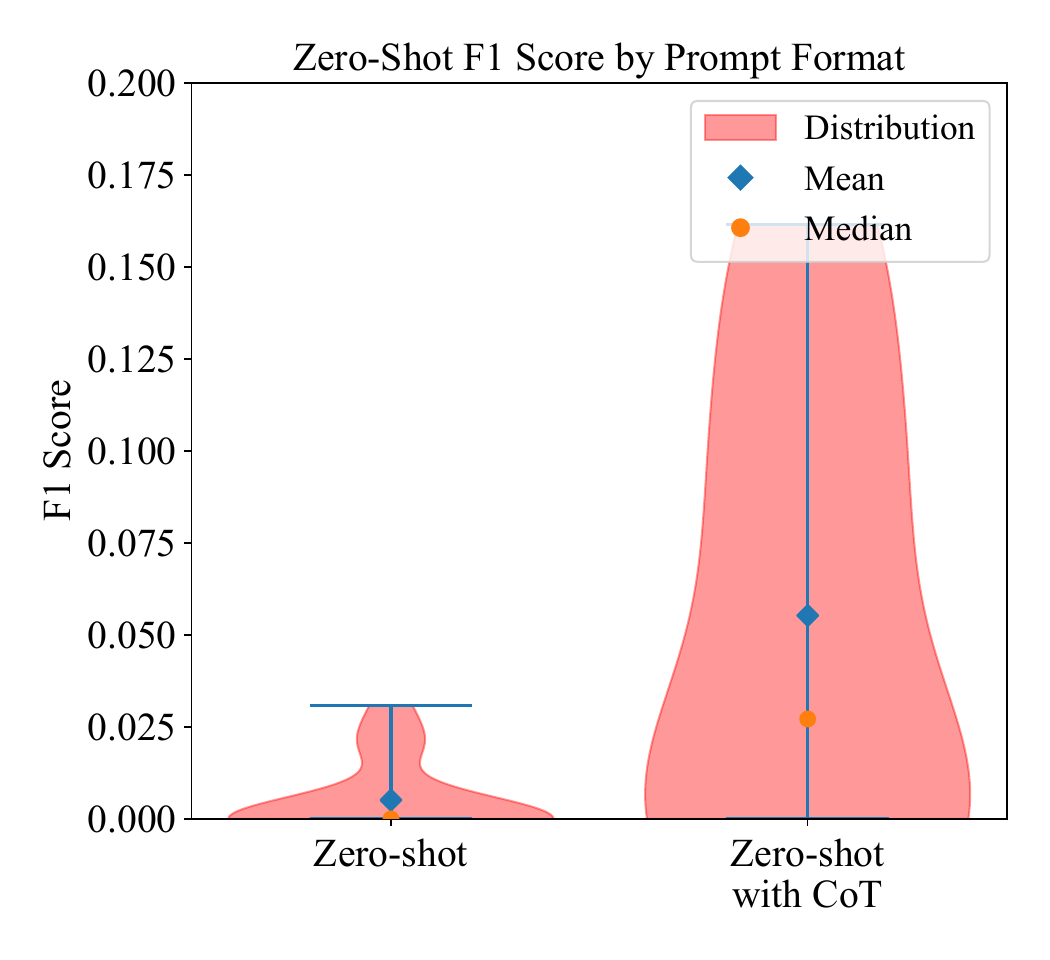}
\caption{Prompt format, F1 score}
\end{subfigure}
\hspace{5mm}
\begin{subfigure}[t]{0.33\textwidth}
\centering
\includegraphics[width=\linewidth]{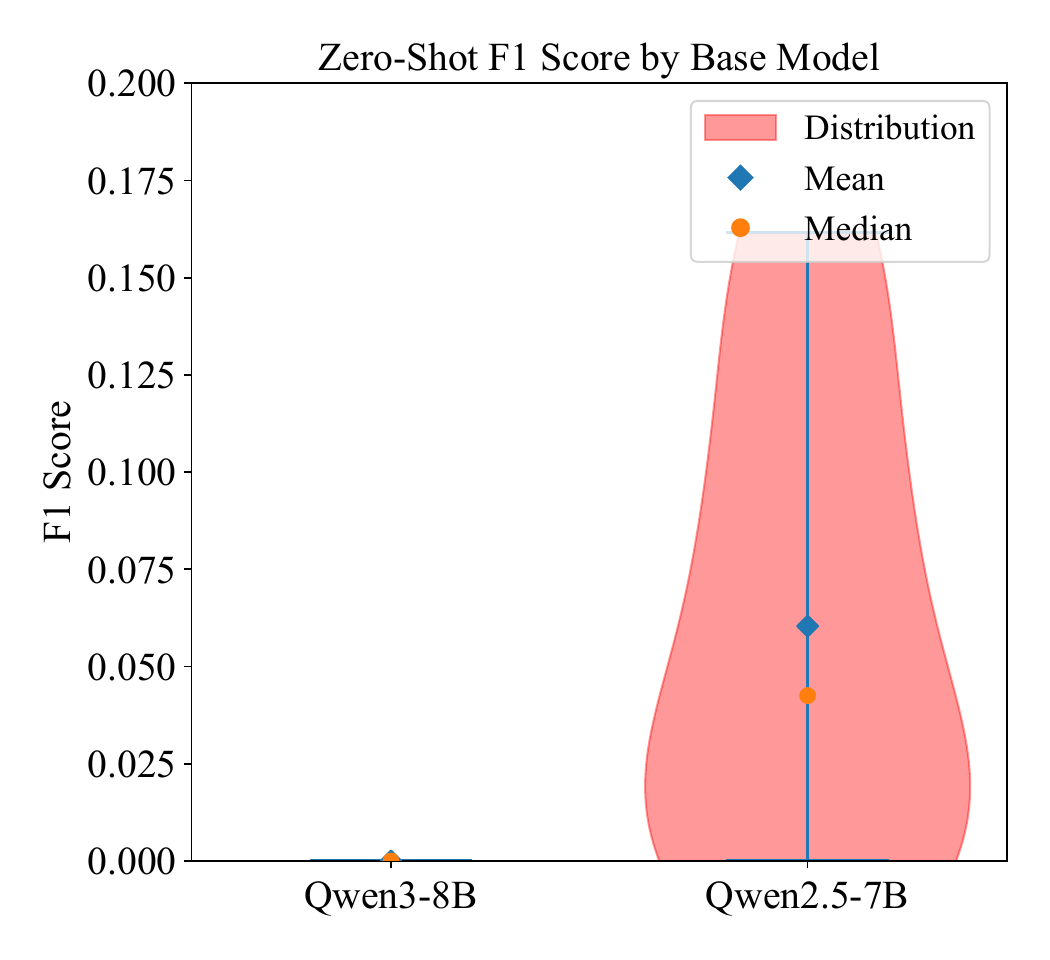}
\caption{Base model, F1 score}
\end{subfigure}

\vspace{0.15em}

\begin{subfigure}[t]{0.33\textwidth}
\centering
\includegraphics[width=\linewidth]{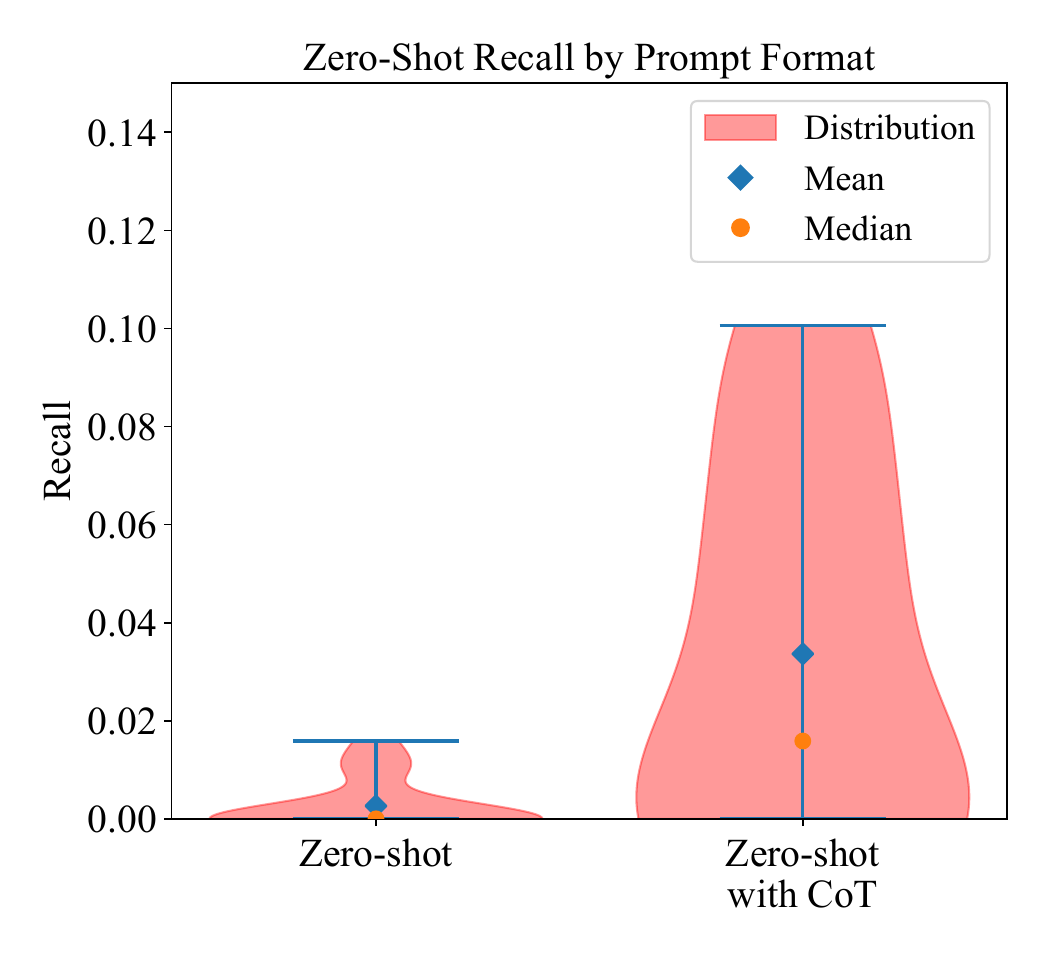}
\caption{Prompt format, Recall}
\end{subfigure}
\hspace{5mm}
\begin{subfigure}[t]{0.33\textwidth}
\centering
\includegraphics[width=\linewidth]{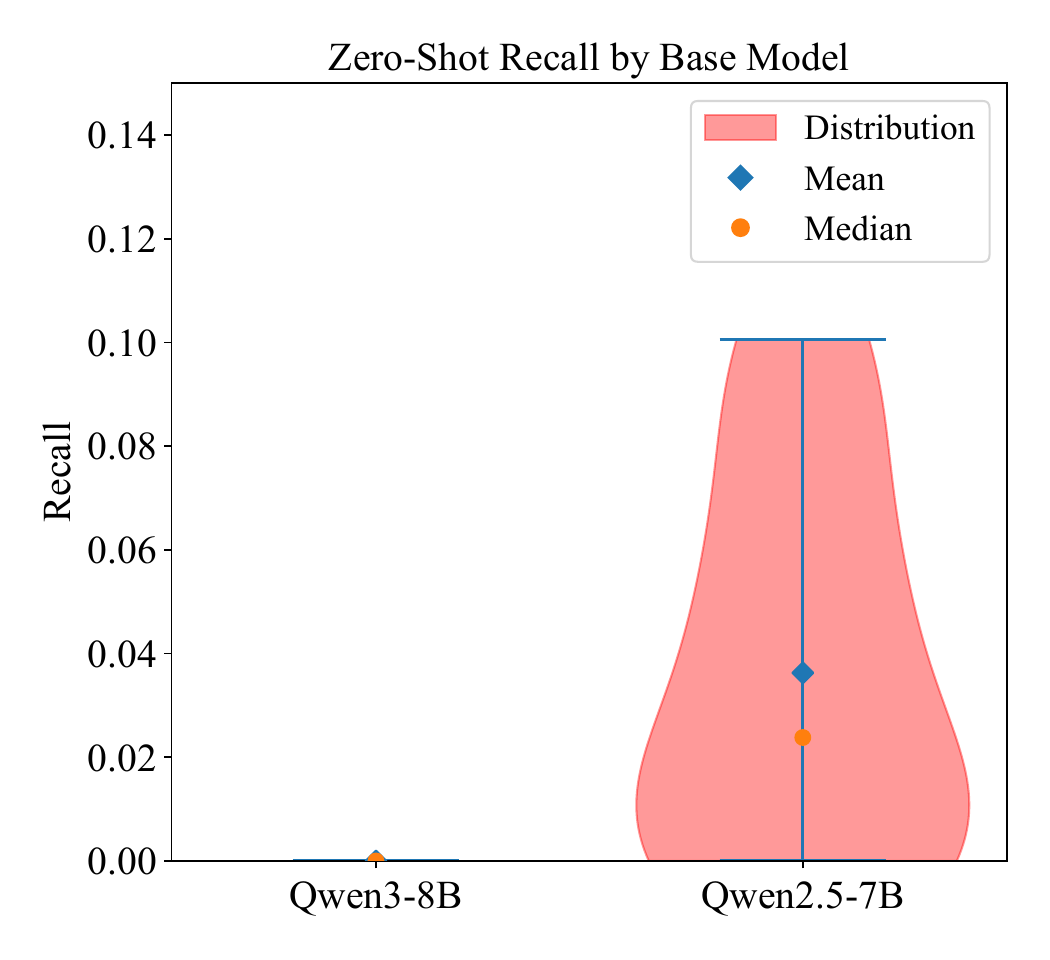}
\caption{Base model, Recall}
\end{subfigure}
\caption{Zero-shot baseline performance by prompt format and by base model.
Panels A--B provide the ROC-AUC reference points for the few-shot ablation;
Panels C--H show the complementary PR-AUC, F1 score, and recall results.}
\label{fig:app-zero-shot-ranking}
\label{fig:app-zero-shot-model-thresholded}
\end{figure*}

For complementary zero-shot metrics, zero-shot with CoT did not provide a uniform improvement over the zero-shot prompt. Across matched model and decoding settings, mean PR-AUC was lower for zero-shot with CoT than for zero-shot prompting (0.351 versus 0.391; paired $t$-test, $P=0.0134$). At the fixed 0.5 decision threshold, zero-shot with CoT increased mean F1 score from 0.005 to 0.055 ($P=0.0084$) and mean recall from 0.0026 to 0.0336 ($P=0.0084$), although the absolute recall values remained low.
Figure~\ref{fig:app-zero-shot-ranking} reports these zero-shot prompt and base-model comparisons for ROC-AUC, PR-AUC, F1 score, and recall.

The Qwen2.5-7B-Instruct subgroup showed why the non-ROC metrics are important for interpreting the zero-shot with CoT effect. Within Qwen2.5-7B-Instruct, zero-shot with CoT increased mean ROC-AUC from 0.538 to 0.565 across matched decoding settings (paired $t$-test, $P=0.041$), but decreased mean PR-AUC from 0.383 to 0.338 ($P=0.0168$). F1 score and recall moved in the opposite direction from PR-AUC because zero-shot with CoT produced more positive predictions at the fixed 0.5 threshold: mean F1 score increased from 0.010 to 0.111 and mean recall increased from 0.005 to 0.067. Thus, the Qwen2.5-7B-Instruct result should not be interpreted as a uniform improvement from zero-shot with CoT; it improved broad ranking by ROC-AUC while reducing precision-recall performance and shifting the fixed-threshold operating point.

\begin{figure*}[!p]
\centering
\begin{subfigure}[t]{0.38\textwidth}
\centering
\includegraphics[width=\linewidth]{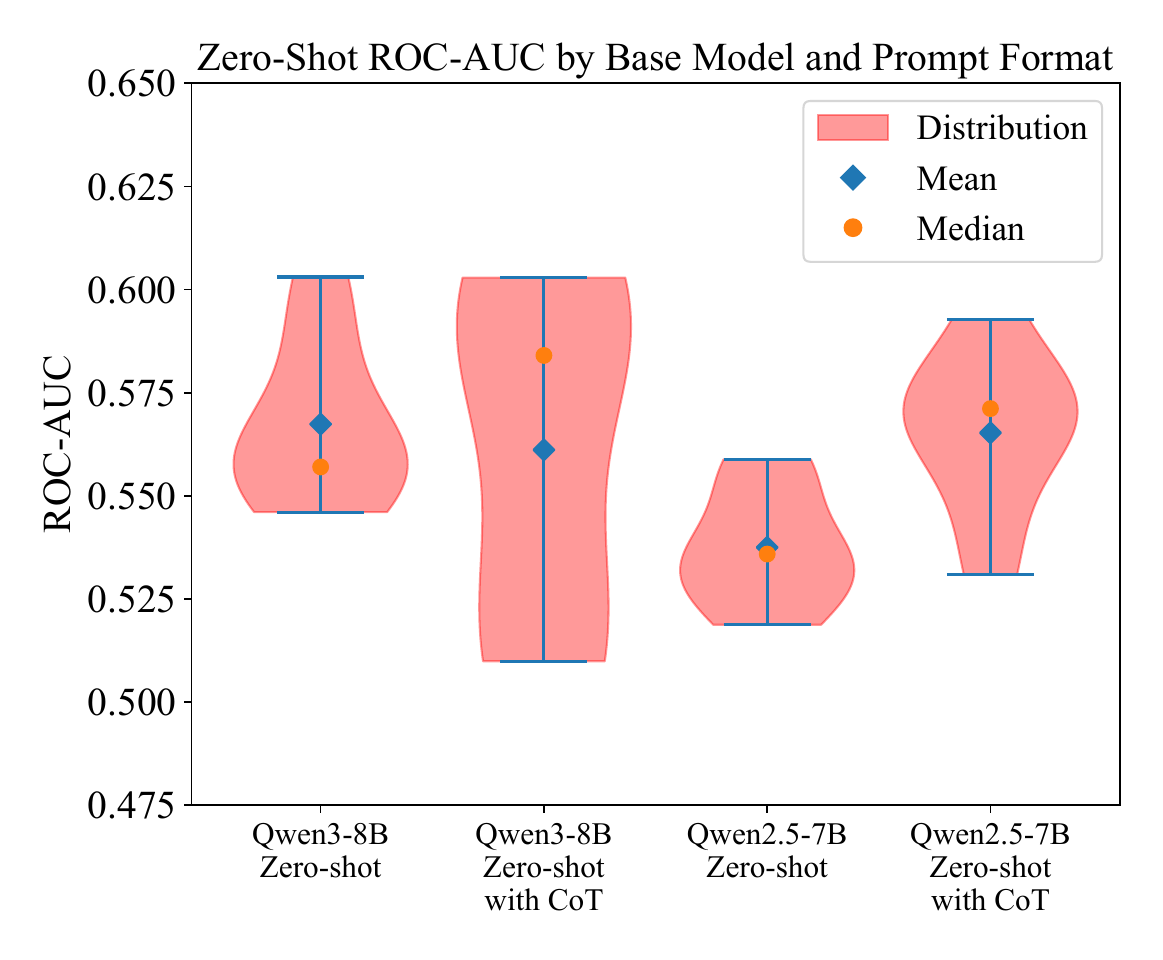}
\caption{Base model and prompt format, ROC-AUC}
\end{subfigure}
\hspace{5mm}
\begin{subfigure}[t]{0.38\textwidth}
\centering
\includegraphics[width=\linewidth]{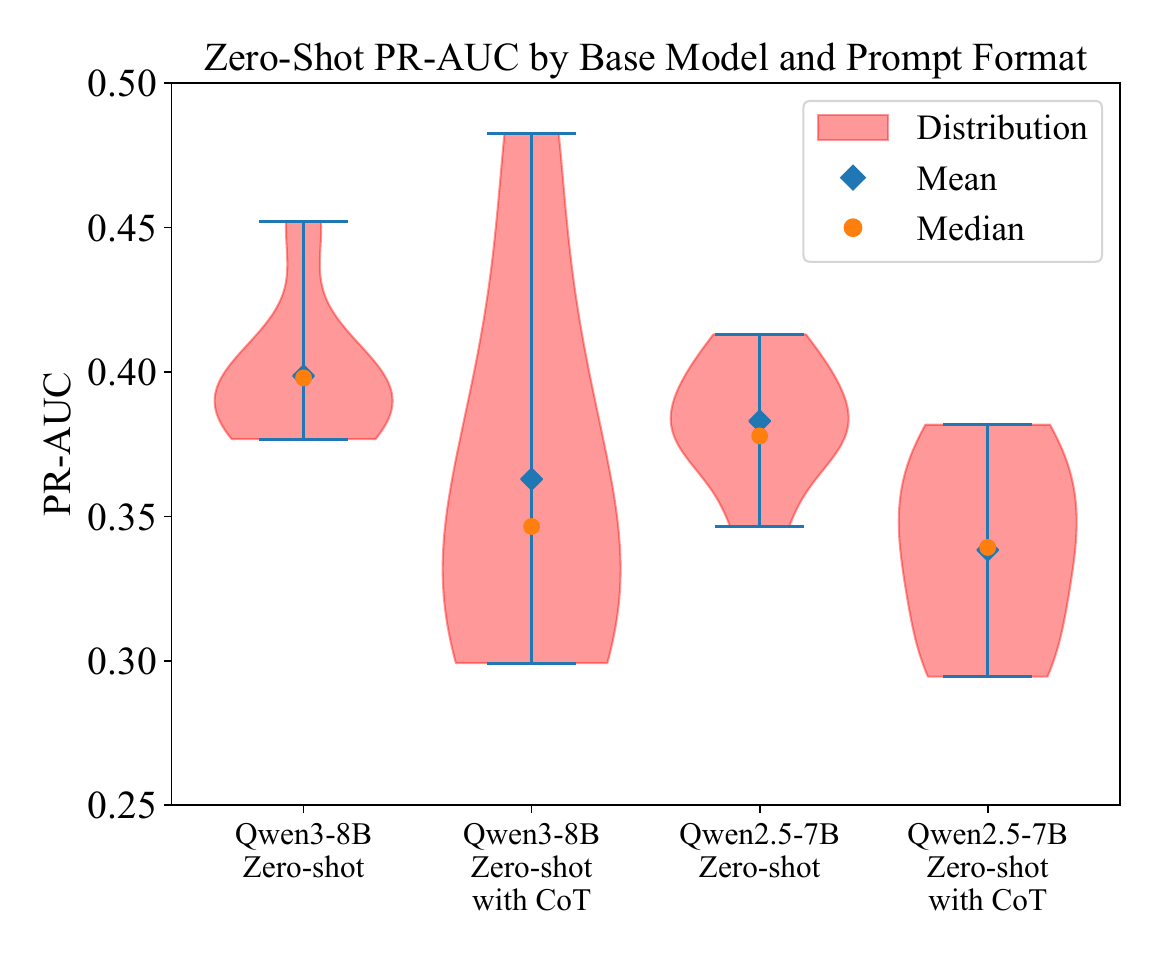}
\caption{Base model and prompt format, PR-AUC}
\end{subfigure}

\vspace{0.5em}

\begin{subfigure}[t]{0.38\textwidth}
\centering
\includegraphics[width=\linewidth]{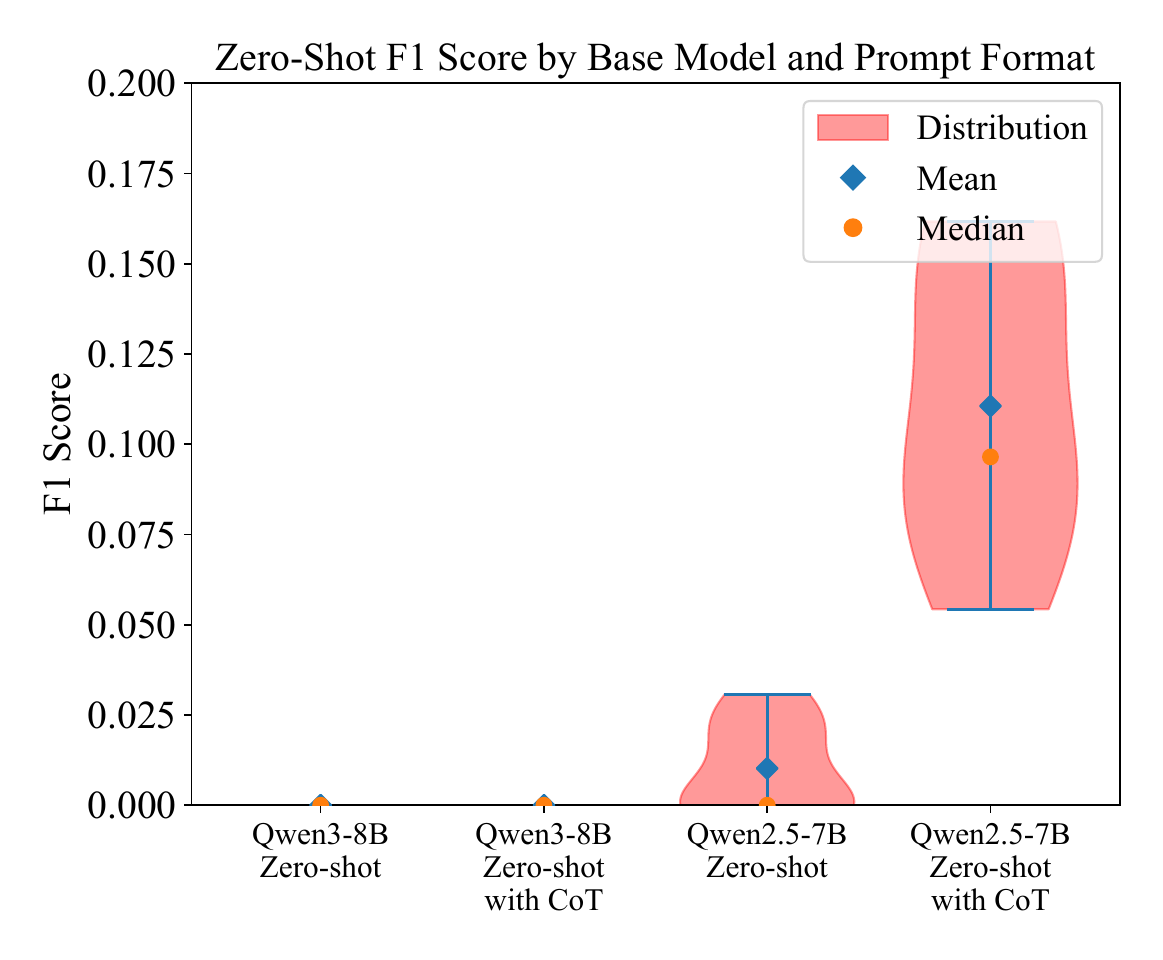}
\caption{Base model and prompt format, F1 score}
\end{subfigure}
\hspace{5mm}
\begin{subfigure}[t]{0.38\textwidth}
\centering
\includegraphics[width=\linewidth]{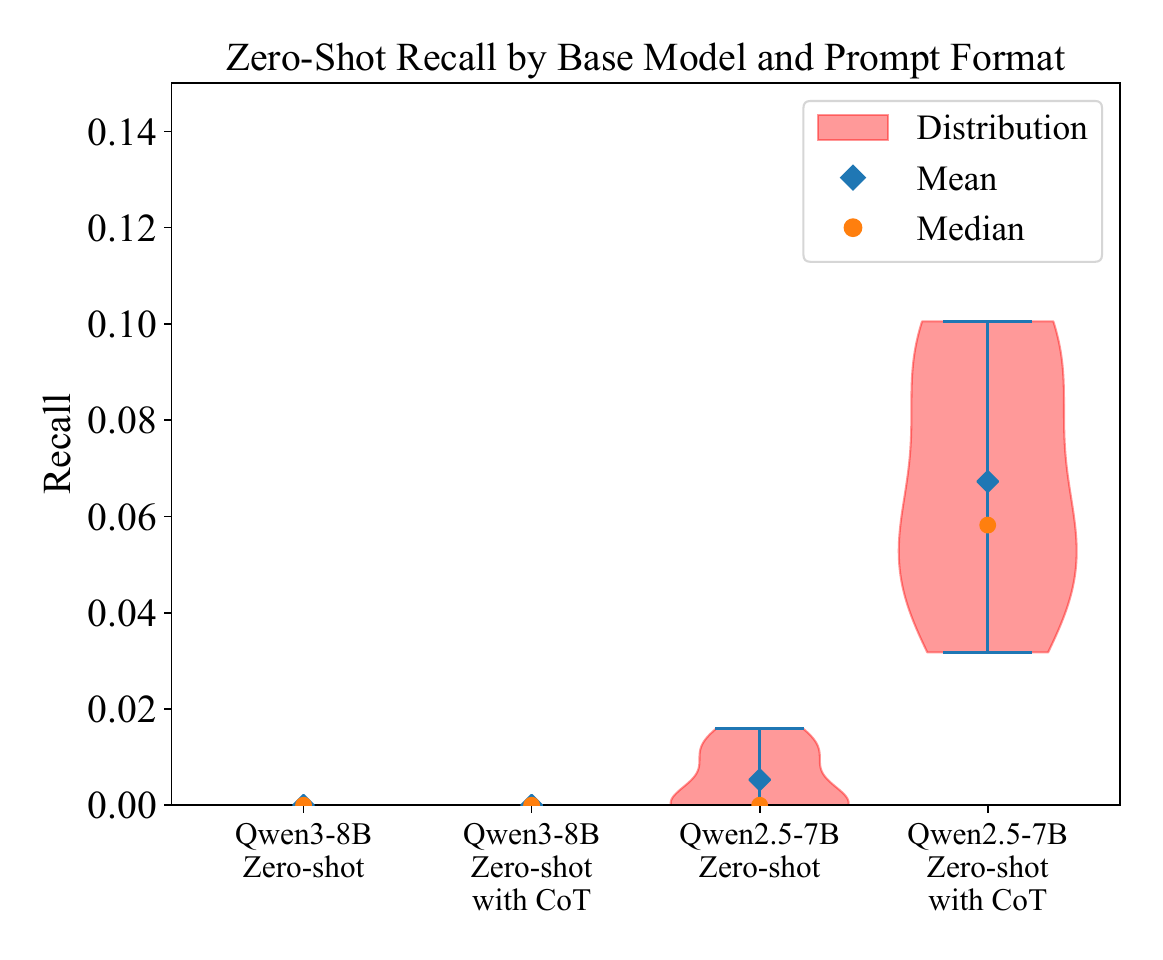}
\caption{Base model and prompt format, Recall}
\end{subfigure}
\caption{Zero-shot base-model-by-prompt-format interaction panels for ROC-AUC, PR-AUC, F1 score, and Recall.}
\label{fig:app-zero-shot-model-prompt-interaction}
\end{figure*}

For the base-model comparison, Qwen3-8B did not significantly improve PR-AUC over Qwen2.5-7B-Instruct across matched zero-shot settings (0.381 versus 0.361; paired $t$-test, $P=0.209$). Qwen3-8B had lower F1 score and recall than Qwen2.5-7B-Instruct (F1 score 0.000 versus 0.060, $P=0.0023$; recall 0.000 versus 0.036, $P=0.0029$). This thresholded limitation was driven by the score scale rather than by the complete absence of ranking signal: Qwen3-8B produced only 0--2 predictions at or above 0.5 per setting, all of them false positives, while all ADRD-case scores stayed below 0.5.
Figure~\ref{fig:app-zero-shot-model-prompt-interaction} breaks the zero-shot results down by the crossing of base model and prompt format, showing that the apparent prompt effect depends on both the metric and the model.

\end{document}